%% file: aui.tex
\documentclass[12pt, final]{elsarticle}

\usepackage{amssymb}

\usepackage{lineno}

\usepackage[margin=1in]{geometry}
\makeatletter
\def\ps@pprintTitle{%
 \let\@oddhead\@empty
 \let\@evenhead\@empty
 \let\@oddfoot\@empty 
 \let\@evenfoot\@empty}
\makeatother

\journal{}

\usepackage{algorithm,algorithmic}
\usepackage[caption=false,font=footnotesize]{subfig}
\usepackage{hyperref}
\usepackage{booktabs}
\usepackage{comment}

\usepackage{xcolor}

\begin{document}

\begin{frontmatter}



\title{Acting upon Imagination: When to Trust Imagined Trajectories in Model Based Reinforcement Learning}

\author[label1]{Adrian~Remonda}
\author[label1,label2]{Eduardo~Veas}
\author[label2]{Granit~Luzhnica}


\affiliation[label1]{organization={Know-Center},
            city={Graz},
            country={Austria}}

\affiliation[label2]{organization={Graz University of Technology},
            city={Graz},
            country={Austria}}

\begin{abstract}
	Model-based reinforcement learning (MBRL) aims to learn model(s) of the environment dynamics that can predict the outcome of its actions. Forward application of the model yields so called imagined trajectories (sequences of action, predicted state-reward) used to optimize the set of candidate actions that maximize expected reward. The outcome, an ideal imagined trajectory or plan, is imperfect and typically MBRL relies on model predictive control (MPC) to overcome this by continuously re-planning from scratch, incurring thus major computational cost and increasing complexity in tasks with longer receding horizon.

	We propose uncertainty estimation methods for online evaluation of imagined trajectories to assess whether further planned actions can be trusted to deliver acceptable reward.
	These methods include comparing the error after performing the last action with the standard expected error and using model uncertainty to assess the deviation from expected outcomes. Additionally, we introduce methods that exploit the forward propagation of the dynamics model to evaluate if the remainder of the plan aligns with expected results and assess the remainder of the plan in terms of the expected reward.
	Our experiments demonstrate the effectiveness of the proposed uncertainty estimation methods by applying them to avoid unnecessary trajectory replanning in a shooting MBRL setting. Results highlight significant reduction on computational costs without sacrificing performance.
	\end{abstract}


\end{frontmatter}


\input{sections/intro.tex}

\input{sections/rw.tex}
\input{sections/methods.tex}

\input{sections/experiments.tex}

\input{sections/discussion}

\label{}

\section*{Acknowledgment}
Funding in direct support of this work: Adrian Remonda reports financial support was provided by AVL List GmbH. Adrian Remonda reports a relationship with Know-Center GmbH that includes: employment. This research was partially funded by AVL GmbH and Know-Center GmbH. Know-Center is funded within the Austrian COMET Program-Competence Centers for Excellent Technologies - under the auspices of the Austrian Federal Ministry of Transport, Innovation and Technology, the Austrian Federal Ministry of Economy, Family and Youth and by the State of Styria. COMET is managed by the Austrian Research Promotion Agency FFG.





\bibliographystyle{plainnat}

\bibliography{references}



\newpage
\appendix
\noindent
\input{sections/appendices}

\end{document}

%% file: sections/intro.tex
\section{Introduction}
Reinforcement learning can be successfully applied in continuous control of complex and highly non-linear systems. Algorithms for reinforcement learning can be categorized as model free (MFRL) or model based (MBRL).
Using deep learning in MFRL has achieved success in learning complex policies from raw input, such as solving problems with high-dimensional state spaces ~\citep{Sutton:1998:IRL:551283}~\citep{mnih2015humanlevel} and continuous action space optimization problems with algorithms like deep deterministic policy gradient (DDPG)~\citep{DBLP:journals/corr/LillicrapHPHETS15}, Proximal Policy Optimization (PPO) ~\citep{schulman2017proximal} or Soft Actor-Critic (SAC) ~\citep{haarnoja2018soft}.
While significant progress has been made, the high sample complexity of MFRL limits its real-world applicability.
Collecting millions of transitions needed to converge is not always feasible in real-world problems, as excessive data collection is expensive, tedious or can lead to physical damage \citep{williams2017information,kurtl2018deep}. Instead, model based methods are comparably sample efficient. MBRL techniques build a predictive model of the world to imagine trajectories of future states and plan the best actions to maximize a reward.

MBRL uses a dynamics model to predict the outcome of taking an action in an environment with given states.
It can bootstrap from existing experiences and is versatile to changing objectives on the fly. Nevertheless, its performance degrades with growing model error.
In the general case of nonlinear dynamics guarantees of local optimality do not hold\citep{Janner2019}.
Sampling-based MPC (model predictive control) algorithms can be used to address this issue. A Sampling-based MPC generates a large number of trajectories with the goal of maximizing the expected accumulated reward. But, complex environments are typically partially observable and the problem is formulated as a receding horizon planning task. Hence, after executing a single step, trajectories are generated again from scratch to deal with the receding horizon and reduce the impact of model compound error \citep{Rao2010}.
An additional challenge lies in the high cost incurred by frequently generating imagined trajectories from scratch during planning. For each trajectory generated, a sequence of actions has to be evaluated with the dynamics model. Each successive step in a trajectory depends on previous states. So the evaluation of a single trajectory is a recurrent process that cannot be parallelized. \textit{Acting upon imagination}, the method here proposed, seeks to estimate online the uncertainty of the imagined trajectory and to reduce the planning cost by continuing to act upon an it unless it cannot be trusted.

Therefore, in this work, we present methods for uncertainty estimation, designed to address and improve the computational limitations of shooting MPC MBRL methods. Our main objective 
is the online estimation of uncertainty of the model plan. A second objective is the application of uncertainty estimation to avoid frequent replanning. Here, we propose using the degree to which computations are reduced as a practical and quantifiable proxy for the effectiveness of uncertainty estimation methods. If our methods are successful, we expect to observe a significant decrease in computations without a substantial decay in performance. The balance between computational efficiency and performance demonstrates the reliability of our uncertainty estimation methods.
A robust estimation of uncertainty facilitates efficient decision-making and optimizes the use of computational resources. 
Our contributions are as follows:

\begin{itemize}
    \item We provide a thorough analysis and discussion on  quantifying the error of predicted trajectories.
    \item We propose methods for uncertainty estimation in MBRL:
    \begin{itemize}
        \item Methods that observe the outcome of the last action in the trajectory i) comparing the error after performing the last action with the standard expected error, ii) assessing the deviation with respect to expected outcomes using model uncertainty.
        \item Methods that exploit the forward propagation of the dynamics model to iii) evaluate if the remainder of the plan aligns with expected results, iv) assess the remainder of the plan in terms of the expected reward.
    \end{itemize}
    \item We demonstrate how our proposed uncertainty estimation methods can be used to bypass the need for replanning in sampling-based MBRL methods.
\end{itemize}

Our experimental results on challenging benchmark control tasks demonstrate that the proposed methods effectively leverage accurate predictions as well as dynamically decide when to replan. This approach leads to substantial reductions in training time and promotes more efficient use of computational resources by eliminating unnecessary replanning steps.

%% file: sections/rw.tex
\section{Related work}

Model-based reinforcement learning (MBRL) has been applied in various real-world control tasks, such as robotics. Compared to model-free approaches, MBRL tends to be more sample-efficient~\citep{Deisenroth2013}.
MBRL can be grouped into four main categories~\citep{Zhu2020}:\\
1) Dyna-style algorithms optimize policies using samples from a learned world model ~\citep{Sutton1990IntegratedAF}.\\
2) Model-augmented value expansion methods, such as MVE ~\citep{Oh2017ValuePN}, use model-based rollouts to enhance targets for model-free Temporal Difference updates.\\
3) Analytic-gradient methods can be used when a differentiable world model is available, which adjust the policy through gradients that flow through the model. When compared to traditional planning algorithms that create numerous rollouts to choose the optimal action sequence, analytic-gradient methods are more computationally efficient. Stochastic Value Gradients (SVG) ~\citep{heess2015learning} provide a new way to calculate analytic value gradients using a generic differentiable world model. Dreamer ~\citep{hafner2020dream}, a milestone in the realm of analytic-gradient model-based RL, demonstrates superior performance in visual control tasks. Dreamer expands upon SVG by facilitating the generation of imaginary rollouts within the latent space\\
4) Model Predictive Control (MPC) and sampling-based shooting methods employ planning to select actions. They are notably effective for addressing real-world scenarios since excessive data collection is not only costly and tedious, but it can also result in physical damage. Additionally, sampling-based MPC methods have the capacity to bootstrap from existing experiences and rapidly adapt to changing objectives on the fly. However, a significant drawback to these approaches is their computationally intensive nature ~\citep{Rao2010, kurtl2018deep}. The present work belongs into the latter category.

Recently, it was demonstrated that parametric function approximators, neural networks (NN), efficiently reduce sample complexity in problems with high-dimensional non-linear dynamics~\citep{Nagabandi2018}.
Random shooting methods artificially generate large number of actions~\citep{Rao2010} and MPC is used to select candidate actions~\citep{camacho2004MPC}. I.e.~\cite{williams2017information} and \cite{DBLP:articles/Drews2017} introduced a sampling-based MPC with dynamics model to sample a large number of trajectories in parallel. A two-layer NN trained from maneuvers performed by a human pilot was superior compared with a physics model built using vehicle dynamics equations from bicycle models. One disadvantage is that NNs cannot quantify predictive uncertainty.

\citet{lakshminarayanan2016simple} utilized the ensembles of probabilistic NNs to determine predictive uncertainty. \citet{Kalweit2017} used a notion of uncertainty within a model-free RL (MFRL) agent to switch executing imagined trajectories from a dynamics model when MFRL agent has a high uncertainty. Conversely, \citet{Buckman2018} used imagined trajectories to improve the sample complexity of an MFRL agent. They improved the Q function by using ensembles to estimate uncertainty and prioritize trajectories thereupon.

Measuring the reliability of a learned dynamics model when generating imagined trajectories has been proposed in several works.
\citet{kurtl2018deep} identified two types of uncertainty: aleatoric (inherent to the process) and epistemic (resulting from datasets with too few data points). The former is the uncertainty inherent to the process, and the latter results from datasets with too few data points.
They combined uncertainty aware probabilistic ensembles in the trajectory sampling of the MPC with a cross entropy controller and achieved asymptotic performance comparable to Proximal Policy Optimization (PPO) ~\cite{schulman2017proximal} or Soft Actor-Critic (SAC) ~\cite{haarnoja2018soft} with more sample efficient convergence. \citet{Janner2019} generated (truncated) short trajectories with a probabilistic ensemble to train the policy of a MFRL agent, thus improving significantly its sampling efficiency. \citet{yu2020mopo} also exploits the uncertainty of the dynamics model to improve policy learning on an offline RL setting. They learn policies entirely from a large batch of previously collected data with rewards artificially penalized by the uncertainty of the dynamics. These works focus on sample efficiency and improving the performance, our work proposes novel methods to estimate the uncertainty of the dynamics model to determine when to replan.

The authors in ~\citep{DBLP:journals/corr/abs-2004-08830} propose a analytic-gradient-based method that considers the reliability of the learned dynamics model used for imagining the future. They evaluate their approach in the context of enhancing vision-based robotic grasping, aiming to improve sample efficiency in sparse reward environments. In contrast to their method, ours does not require the use of numerous local dynamics models or a self-organizing map. Instead, we introduce a technique that exploits the uncertainty of the dynamics model to estimate the uncertainty of plan during execution, primarily aimed at minimizing replanning within an MPC framework. Close to our work, ~\cite{Zhu2020} studied the discrepancy between imagination. Their method allows for policy generalization to real-world interactions by optimizing the mutual information between imagined and real trajectories, while simultaneously refining the policy based on the imagined trajectories. However, their focus is on analytic gradients MBRL only, our method can be applied to any MBRL which yields a notion of uncertainty and we focus on shooting methods, which are still the first choice in domains like self driving cars \citep{williams2017information}.

~\cite{hansen_temporal_2022} obtained state of the art performance in terms of reward and training time on diverse continuous control tasks by significantly improving Model-Augmented Value Expansion methods.  Their approach effectively combines the strengths of both MFRL and MBRL. They adopt a learned task-oriented latent dynamics model for localized trajectory optimization over a short horizon. Furthermore, they utilize a learned terminal value function to estimate the long-term returns. However, their method still necessitates the learning of the value function. Depending on the context, this could present challenges when compared to shooting methods.

Nevertheless, shooting MPC methods still suffer from expensive computation~\cite{kurtl2018deep,Zhu2020}.
Thus, our research seeks to reduce the amount of computation continuing to act upon trajectories that seem trustworthy. Our solution builds upon results of~\cite{kurtl2018deep}, using probabilistic ensembles and cross entropy in the MPC.

%% file: sections/methods.tex
\section{Preliminaries}
RL aims to learn a policy that maximizes the accumulated reward obtained from the environment. At each time \(t\), the agent is at a state \(s_t\in S\), executes an action \(a_t\in A\) and receives from the environment a reward \(r_t = r(s_t, a_t)\) and a state \(s_{t+1}\) according to some unknown dynamics function \(f:S\times A \to S\). The goal is then to maximize the sum of discounted rewards \(\sum_{i=t}^\infty \gamma^{(i-t)} r(s_i, a_i)\), where \(\gamma\ \in [0,1]\).
MBRL uses a discrete time dynamics model \(\hat{f}=(s_t, a_t)\) to predict the future state \(\hat{s}_{t+\Delta_t}\) after executing action \(a_t\) at state \(s_{t}\). To reach a state into the future, the dynamics model evaluates sequences of actions, $a_{t:t+H}=(a_t, \ldots, a_{t+H-1})$ over a longer horizon \(H\), to maximize their discounted reward \(\sum_{i=t}^{t+H-1} \gamma^{(i-t)} r(s_i, a_i)\).
Due to partial observability of the environment and the error of the dynamics model \(\hat{f}\) in predicting the real physics \(f\), the controller typically executes only one action \(a_t\) in the trajectory and the optimization is solved again with the updated state \(s_{t+1}\).
Algorithm~\ref{alg:algorithm} outlines the general steps.
When training from scratch, the dynamics model $\hat{f}_{\theta}$ is learned with data, $\mathcal{D}_{env}$, collected on the fly. With $\hat{f}_{\theta}$, the simulator starts and the controller is called to plan the best trajectory resulting in $a^*_{t:t+H}$. Only the first action of the trajectory $a_t^*$ is executed in the environment and the rest is discarded. The data collected from the environment is added to $\mathcal{D}_{env}$ and $\hat{f}_{\theta}$ is trained further. MBRL requires a strategy to generate an action \(a_t\), given a state \(s_t\), a  discrete time dynamics model \(\hat{f}=(s_t, a_t)\) to predict the state \(s_{t+1}\), and a reward function \(r_t = r(s_t, a_t)\).

\paragraph{Probabilistic Dynamics Model}
We model the probability distribution of next state given current state and an action using a neural network based regression model similar to~\cite{lakshminarayanan2016simple}. The last layer of the model outputs parameters of a Gaussian distribution modeling the aleatoric uncertainty (due to the randomness of the environment). Its parameters are learned together with the parameters of the neural network. To model the epistemic uncertainty (of the dynamics model due to generalization errors), we use ensembles with bagging where all the members of the ensemble are identical except for their initial weight values.
Each ensemble element takes as input the current state and action $s_t$ and $a_t$, and it is trained to predict the difference between $s_t$ and $s_{t+1}$, instead of directly predicting the next step. Thus, the learning objective for the dynamics model becomes, $\Delta s = s_{t+1} - s_{t}$.
$\hat{f}_{\theta}$ outputs the probability distribution of the future state $p_{s(t+1)}$ from which we can sample the future step and its confidence $\hat{s}, \hat{s}_{\sigma} = \hat{f}_\theta(s, [\textbf{a}])$, where $s_{\sigma}$ captures both, epistemic and aleatoric uncertainty.

\begin{algorithm}
	\caption{MBRL}
	\label{alg:algorithm}
	\small
	\begin{algorithmic}[1] 
	    \STATE {Set replay buffer $\mathcal{D}$ with one iteration of random controller}
		\FOR{Iteration $i = 1$ \TO $NIterations$}
		\STATE Train $\hat{f}$ given $\mathcal{D}$
		\FOR{Time $t = 0$ \TO $TaskHorizon$}
		\STATE Get $a_{t:t+H}^{*}$ from $CompOptTrajectory(s_{t},\hat{f})$

		\STATE Execute first action $a_t^{*}$ from optimal actions $a_{t:t+H}^{*}$
		\STATE Record outcome: $\mathcal{D} \leftarrow \mathcal{D} \cup \{s_t,a_t^{*},s_{t+1}\}$

		\ENDFOR
		\ENDFOR
	\end{algorithmic}
\end{algorithm}
\paragraph{Trajectory Generation}
Each ensemble element outputs the parameters of a normal distribution. To generate trajectories, P particles are created from the current state, $s_t^p=s_t$, which are then propagated by: $s_{t+1}^p \sim \hat{f}_b(s_{t}^p, a_{t})$, using a particular bootstrap element $b \in \{1,...,B\}$. There are many options on how to propagate the particles through the ensemble as analyzed in detail in ~\cite{kurtl2018deep}.
They obtained the best results using the $TS_\infty$ method, which refers to particles never changing the initial bootstrap element. Doing so, results in having both uncertainties separated at the end of the trajectory. Specifically, aleatoric state variance is the average variance of particles of same bootstrap, whilst epistemic state variance is the variance of the average of particles of same bootstrap indexes. Our approach also uses the $TS_\infty$ method.

\paragraph{Planning}
To select a convenient course of action leading to \(s_H\), MBRL generates a large number of trajectories $K$ and evaluates them in terms of reward.
To find the actions that maximize reward,
we used the cross entropy method (CEM) ~\cite{BOTEV201335}, which is an algorithm for solving optimization problems based on cross-entropy minimization. CEM gradually changes the sampling distribution of the random search so that the rare-event is more likely to occur. Thus, this method estimates a sequence of sampling distributions that converges to a distribution with probability mass concentrated in a region of near-optimal solutions.
Algorithm~\ref{alg:optimal_trajectory} describes the use CEM to compute the optimal sequence of actions $a^*_{t:t+H}$.
The controller uses a single action in a trajectory, the computational complexity is constant at each step, given by the depth of the task horizon ($H$) and the number of trajectories ($K$) or breadth. It is possible to parallelize in breadth, but the evaluation of some action \(a_{t+i}\) at state \(s_{t+i}\) with dynamics model \(\hat{f}\) is iterative, requiring knowledge of at least one past state and cannot be parallelized in depth. This leads to complexity O(H  x  A  x  K), where A refers to the dimension of actions (how many controllable aspects). $A$ and $K$ depend on the environment.

\begin{algorithm}
	\caption{Compute Optimal Trajectory - Planning}
	\label{alg:optimal_trajectory}
	\small
	\textbf{Input}: $s_{init}$: current state of the environment, dyn. model $\hat{f}$
	\begin{algorithmic}[1] 
		\STATE Initialize $P$ particles, $s_{\tau}^p$, with the initial state, $s_{init}$
		\FOR{actions sampled $a_{t:t+H}$ in $1 \dots CEM_{samples}$}
		\STATE Propagate state particles $s_{\tau}^p$ using TS and $\hat{f}|\{\mathcal{D}, a_{t:t+H}\}$
		\STATE Evaluate actions as $\sum\limits_{\tau=t}^{t+H}{\frac{1}{P} \sum\limits_{p=1}^{P} r(s_{\tau}^p, a_{\tau}) }$
		\STATE Update CEM$(.)$ distribution
		\ENDFOR
	\STATE \textbf{return} $a^{*}_{t:t+H}$
	\end{algorithmic}
\end{algorithm}

\section{The Promise of Imagination}

Generating trajectories is an essential part of the entire process. The predicted states within those trajectories may have a high variance and their quality will depend on the complexity of the environment as well as the number of steps in the future, H. Estimation and online update of uncertainty is needed to determine if a trajectory is reliable. We contend that when predicted trajectories are \textit{reliable} it is not necessary to frequently replan them.

Starting from a state at time $t$, a run of the planner yields the optimal set of actions $a^*$ of H steps. Using $a^*$, the dynamics model yields the probability of the next state from which we can sample the next state and confidence $\hat{s}_{t+1}, \sigma_{t+1} \sim p^*_{s(t+1)}$  and future reward $\hat{r}_{t} \sim p^*_{r(t+1)}$.
Thus, one step sampled from the imagined trajectory is composed of $\langle s_t, a_t, \hat{s}_{t+1}, \sigma_{t+1}, \hat{r}_{t+1}\rangle$, where $s_t$ represents the current state, $a_t$ the action to be taken, $\hat{r}_{t+1}$ is the predicted reward if $a_t$ is executed, $\hat{s}_{t+1}$ the predicted next state and $\sigma_{t+1}$ the confidence bound issued by the dynamics model for the prediction. Then, the planner generates iteratively the entire trajectory $\tau_t$ of H steps, where each step $i$ is composed of the probability distribution $p^*_{s(t+i)}$.

\begin{figure}[ht]
	\includegraphics[width=\linewidth]{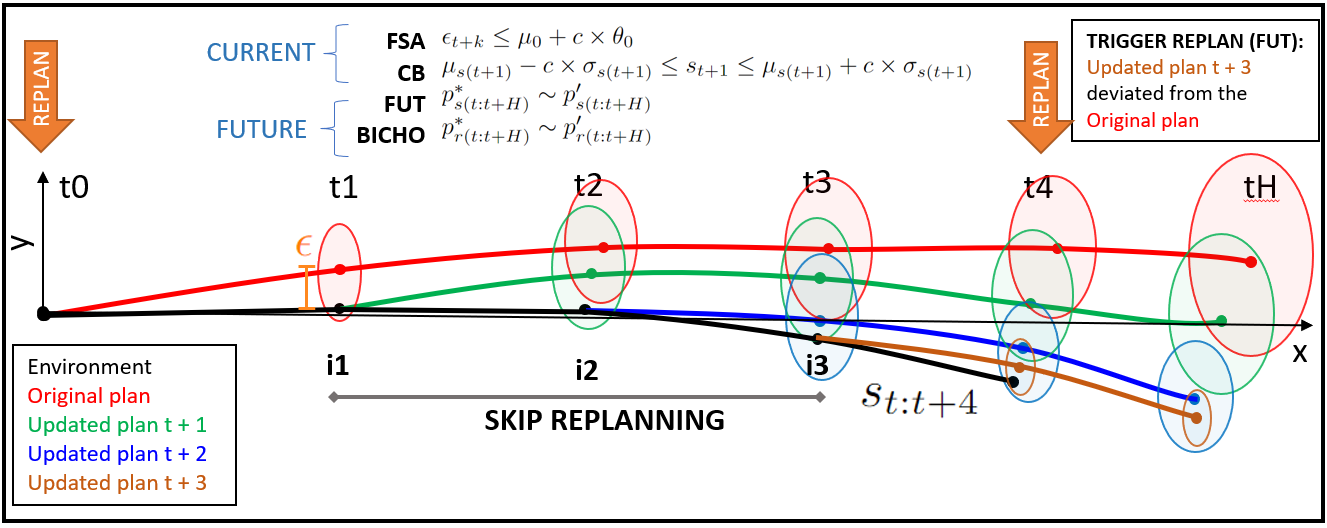}
	\caption{The provided figure depicts a running example for a proposed method called FUT with task horizon H=5 and demonstrates a replanning event at time step t=4. The states obtained from the environment are represented in black.
	The red line represents the imagined trajectory at state $s_t$, while the green, blue and orange lines show the projected outcomes from states $s_{t+1}$, $s_{t+2}$ and $s_{t+3}$, respectively. A replanning was trigger at $s_{t+4}$ as the projected trajectory shown in orange deviates from the expected outcome.\\
	FSA and CB methods observe the outcome of the last action in the trajectory. FSA compares the error after performing the last action with the standard expected error, while CB assesses the deviation with respect to expected outcomes using model uncertainty. ($\epsilon$ in the figure). FUT and BICHO exploit the forward propagation of the dynamics model. FUT evaluates whether the remainder of the plan aligns with expected results, and BICHO assesses the remainder of the plan in terms of the expected reward. \label{fig:running_example}
	}
\end{figure}

Our methods stem from the following information obtained after executing each step in a trajectory: instantaneous deviation between predicted and real outcome, and impact on the projected plan, see Fig.~\ref{fig:running_example}.
Executing the action $a_t$ from the imagined trajectory $\tau_t$, yields a real-world state $s_{t+1}$ and reward $r_{t+1}$. The state $s_{t+1}$ is expected to fall within the uncertainty of the model $\hat{s}_{t+1}, \sigma_{t+1} \sim p^*_{s(t+1)}$. Given \(f(s_t, a_t)=s_{t+1}\) and \(\hat{f}(s_t, a_t)=\hat{s}_{t+1}\), the instantaneous error \(\epsilon_{t}= |s_{t+1}-\hat{s}_{t+1}| \geq 0\) can be measured. The error at step $i$, $\epsilon_{t+i}$ refers to the error after trajectory calculation and executing $a_t$. We can model the error distribution and observe if new errors fall within.
This refers to immediate effect on the last state.
As regards impact on future actions, forward application of the model to the remainder of the actions starting at the new state yields a new trajectory with projected state $p^*_{s(t+1:t+H)}$ and projected reward $p^*_{r(t+1:t+H)}$, which can be compared with the planned expected outcomes.

\begin{figure}[ht] 
	\centering
	\includegraphics[width=\linewidth]{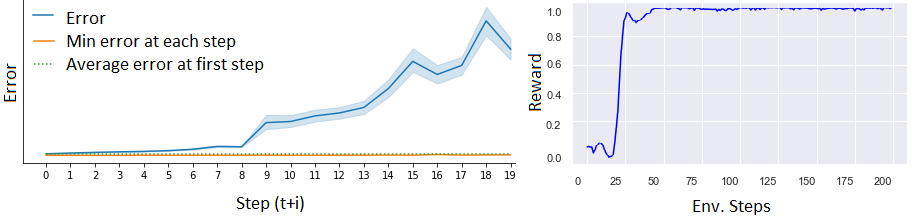}
	\vspace*{-7mm}
	\caption{\textbf{Left} Trajectory errors of predicted future steps along with the minimal error at each step and average error at first step. \textbf{Right} Episode reward of CP as a function of environment steps. The agent task is to hold the pole up, reward ranges from +1 (pole is up) to 0  (pole is down).}
\label{fig:argumentation} 
\end{figure}

\paragraph{Trajectory Quality Analysis}
\label{sec:quality}
As a preliminary experiment, we wanted to analyze the quality of imagined trajectories with a trained dynamics, to determine boundaries of how many actions can be executed without deviating from the plan. We analyzed imagined trajectories on agents in the MuJoCo~\cite{conf/iros/TodorovET12} physics engine, with the Cartpole (CP) environment $S\in \mathbb{R}^{4}, A \in \mathbb{R}^1$, $TaskH$ 200, $H$ 25, with $TaskH$ being the task horizon and $H$ the trajectory horizon. The additional material shows similar findings in other environments.
A dynamics model was pre-trained in conventional MBRL-MPC, running Algorithm~\ref{alg:algorithm} five times from scratch, with trajectory (re-)planning after executing each single action. The best performing model was selected for the analysis. 
The procedure consisted in collecting the errors of predicted and actual state when avoiding the re-planning for $n$ steps.
For each $n\in \{0,...19\}$ the algorithm was run for $TaskH$ steps, $10$ times. Therefore, the error at $n=0$ represents the average error of 10 runs executing the first action in the trajectory.
Figure~\ref{fig:argumentation} illustrates the error of predicted trajectories as a function of $n$ steps used (i.e., avoiding re-planning).
While the error and its variation increases with $n$, the minimum error at each step is still at the same level (and often lower) than the average error at step 0, where re-planning can not be avoided at all. 
Generally, re-planning earlier results in a lower error. However, the chart also shows that some trajectories are so reliable that 19 steps can be executed with error lower than the average error of first point in trajectory.

\textbf{Reward Analysis.}
\label{sec:reward_analysis}
Compared to the vector of state errors, the reward has the advantage of being a more compact representation (single scalar). It also provides substantial information. Figure~\ref{fig:argumentation} right shows the reward of successfully solved task in CP. 
After 50 steps, the reward does not change significantly and the system is at a local equilibrium. We contend that when the system is at equilibrium the dynamics model can reliably anticipate the outcomes of the agent's actions, consequentially rewards are expected to remain similar $\Delta_r(t)=r_t - r_{t-1}\simeq \Delta_r(t+1)=r_{t+1}-r_t$.

\section{Acting Upon Imagination}
From the above discussion, the following information is available after executing each step (shown in Fig~\ref{fig:argumentation}): 
(i) immediate error  ($\epsilon_{t+1}$),  (ii) the model uncertainty or confidence bounds for an action imagined against its execution ($\hat{s}_{t+1}, \sigma_{t+1} \sim p^*_{s(t+1)}$), iii) deviation in projected future states($p^*_{s(t+1:t+H)}$) and iv) the deviation in projected future rewards ($p^*_{r(t+1:t+H)}$). The last two pieces (iii), (iv) are obtained by forward applying the dynamics model with the remainder of the actions starting at the new state.
We leverage each piece of available information to develop methods for uncertainty estimation and evaluate their performance in avoiding replanning events.

Algorithm~\ref{alg:aui_algorithm} presents the core logic of our proposed methods to continue acting upon imagined trajectories and reduce computation. The variable $skip$ is updated with the result of one of four proposed methods in Algorithms~\ref{alg:fac_filter}, ~\ref{alg:cb_filter}, ~\ref{alg:fut_filter} or ~\ref{alg:ccm_filter}. Depending on the outcome, replanning can be avoided and computations reduced.
If $skip$ is False, only the first predicted action is executed in the environment. Otherwise, subsequent actions from $a^*_{t:t+H}$ are executed until the $skip$ flag is set back to False or the $TaskHorizon$ number of steps in the environment is reached.

\begin{algorithm}[t!]
	\caption{Acting Upon Imagination}
	\label{alg:aui_algorithm}
	\small
	\begin{algorithmic}[1] 
		\IF{trainModel}
		\STATE Initialize PE dynamics model $\hat{f}$ parameters
		\STATE {Set replay buffer $\mathcal{D}$ with one iteration of a random controller}
		\ELSE
		\STATE{Load pre trained $\hat{f}$ parameters and replay buffer $\mathcal{D}$}
		\ENDIF
		\STATE $skip = False$
		
		\FOR{Iteration $l = 1$ to $NIterations$}
		
		\IF{trainModel}
		\STATE Train $\hat{f}$ given $\mathcal{D}$
		\ENDIF 
		
		\FOR{Time $t = 0$ to $TaskHorizon$}
		\IF {not $skip$}
		\STATE Get $a_{t:t+H}^{*}$ from $CompOptTrajectory$ $(s_{t}, \hat{f})$ and
		\STATE $p^*_{s(t:t+H)}$, $p^*_{r(t:t+H)}$ given $(s_{t}, \hat{f}, a^*_{t:t+H})$	
		\STATE $i = 0$
		\ELSE
		\STATE $i$ += $1$
		\ENDIF
		\STATE Execute first action $a_t^{*}$ from optimal actions $a_{t:t+H}^{*}$
		\STATE Discard first action and keep the rest $a_t^{*} = a_{t+1:t+H}^{*}$
		\STATE Record outcome: $\mathcal{D} \leftarrow \mathcal{D} \cup \{s_t,a_t^{*},s_{t+1}\}$ 
		\STATE skip = shouldSkip$\{NSKIP|FSA|CB|FUT|BICHO\}$
			\ENDFOR
		\ENDFOR
	\end{algorithmic}
\end{algorithm}

\textbf{N-Skip:} as a baseline, we introduce N-Skip, which is a straight forward method for replanning that executes a fixed $n$ steps in a trajectory (of length $H$) and triggers replanning at step $n+1$ ($n < H$). For $n=0$ the trajectory is recomputed at every step. As earlier replanning generally lead to lower error. $n$ is a hyperparameter that should be tuned to meet the performance requirements. In the CP environment, Figure~\ref{fig:argumentation} left shows a sharply increasing error at $n=7$, which amounts to $88\%$ less computations.
Interestingly, despite its simplicity, N-Skip has not been extensively analyzed or reported in existing literature.

\textbf{First Step Alike (FSA):} Some trajectories are more reliable than others and a cutoff of n skips for all trajectories does not consider this variation in quality. Figure~\ref{fig:argumentation} left shows that there are cases where after 19 steps of a trajectory, the prediction error is still lower than the average error of predicted states right after replanning. To account for such a variation, we propose a dynamic decision making.
We would like to continue acting if the replanning will not improve over the error of predicted states. The error is lowest just after replanning and increases with number of steps. So, the main principle of FSA, is to omit replanning at any point $t+i, i < H$, as long as the error $\epsilon_{t+i-1}$ is comparable to errors right after replanning.\\
In formal terms, assuming a large sample of M errors collected right after replanning ($\epsilon_{0}$) denoted by $\mathbb{E} = \{ \epsilon_{0}^{(m)} \mid m \in \{1...M\}\}$. Actions in a trajectory $\tau_t$ with predicted states $\{ \hat{s}_{t+i+1} \mid i \in \{0...H-1\}\}$ are evaluated at each point $i$, and if the error $\epsilon_{t+i}$ fits the distribution of $\mathbb{E}$, then the replanning is skipped. Otherwise, a new trajectory is generated.
The challenge lies in determining when the error fits the distribution of $\mathbb{E}$. Two methods are proposed for handling this. If the errors $\mathbb{E}$ follow a Gaussian distribution represented by mean $\mu_0$ and standard deviation $\theta_0$, then according to the three sigma rule~\cite{Pukelsheim1994TheTS}, 68.27\%, 95.45\% and 99.73\% of the errors should lie within one, two and three standard deviations of the mean, respectively. It follows that, any given error $\epsilon_{t+i}$ (at point $t+i+1$) such that $ \mu_0 - c \times \theta_0 \leq \epsilon_{t+i}\leq \mu_0 + c \times\theta_0$ fits the distribution and thus the replanning should be skipped. The constant $c$ is a hyper parameter that defines the specificity of such filtering method. 
This filtering ensures that the error $\epsilon_{t+i+1}$ is below a percentile of errors in $\mathbb{E}$ where the percentage number depends on $c$. 
Furthermore, as we do not want to filter out errors that are too small, we could adopt our rule to one side only: $\epsilon_{t+i}\leq \mu_0 + c \times\theta_0$. Finally, for the case where the distribution of $\mathbb{E}$ is not Gaussian, a similar effect can be achieved by ensuring that $\epsilon_{t+i}$ is within the $c$ percentile ($P_{c\%}$) of the errors in $\mathbb{E}$, where $c$ is a parameter to tune. The specific logic for the FSA is given by Algorithm~\ref{alg:fac_filter}.

\begin{algorithm}
	\caption{FSA: deciding whether to skip replanning}
	\label{alg:fac_filter}
	\small
	\textbf{Input}: $\mathbb{E} = \{ \epsilon_{0}^{(i)} \mid i \in \{1...M\}\}$, parameter $c$ and $\epsilon_{t+k}$
	\begin{algorithmic}[1]
		\IF{$\mathbb{E}$ is normally distributed (described by $\mu_{0}$ and $\theta_{0}$)}
		\RETURN TRUE \textbf{if} $\epsilon_{t+k}\leq \mu_{0} + c \times\theta_{0}$ \textbf{else} FALSE 
		\ENDIF
	\RETURN TRUE \textbf{if} $\epsilon_{t+k}\leq P_{c\%}$  \textbf{else} FALSE 

	\end{algorithmic}
\end{algorithm}

\textbf{Confidence Bounds (CB):}
If the dynamics model has a notion of uncertainty, one can obtain a prediction $\hat{s}_{t+1}$ with an uncertainty or confidence $\sigma_{t+1}$.
Given that $p^*_{s(t+1)}$ is modeled by an ensemble of Gaussian regressors, we can assume that the confidence bound $\sigma_{s(t+1)}$ represents the variability in predicted outcomes $\mu_{s(t+1)}$ of an action $a_t$, where the action in question has been deemed appropriate at the given state $s(t)$ therefore is in the trajectory. After performing the action, we obtain a real-world state $s_{t+1}$. This method considers the trajectory reliable, if the actual state $s_{t+1}$ is close to predicted state $\mu_{s(t+1)}$ within the confidence bound $\sigma_{s(t+1)}$ of predicted output states obtained from the dynamics model, meaning:
$\mu_{s(t+1)} - c \times \sigma_{s(t+1)} \leq s_{t+1} \leq \mu_{s(t+1)} + c \times \sigma_{s(t+1)}$, where $c$ is a constant representing the selectivity of the filter adjusted with a factor of sigma, a hyper parameter to be tuned. The specific logic for the CB method is given by Algorithm~\ref{alg:cb_filter}.
In a nutshell, this method assumes that the performance of an action could lead to several \emph{expected} possible outcomes (bounded by the prediction). After performing the action, it is observed whether the outcome lies within the boundary of expected outcomes to determine the reliability of the trajectory.

\begin{algorithm}
	\caption{CB: deciding whether to skip replanning}
	\label{alg:cb_filter}
	\small
	\textbf{Input}: $s_{t+1}$, $p^*_{s(t+1)}$ and $c$ 
	\begin{algorithmic}[1]
		\STATE Get $\mu_{s(t+1)}$, $\sigma_{s(t+1)}$ from  $p^*_{s(t+1)}$
		\RETURN TRUE \textbf{if} ($\mu_{s(t+1)} - c \times \sigma_{s(t+1)} \leq s_{t+1}  \leq \mu_{s(t+1)} + c \times \sigma_{s(t+1)})$ \textbf{else} FALSE
	\end{algorithmic}
\end{algorithm}

\textbf{Probabilistic future trust (FUT):}
FSA and CB asses the error of the state obtained after executing an action vs the prediction estimated by $\hat{f}_{\theta}$. Instead, FUT regards the effect of the last action on the outcomes of future actions, by projecting the remaining imagined actions in the trajectory from the newly obtained state.
After replanning, the trajectory $\tau_{t+1}$ of H steps, where each step $i,i=0...H$ is composed of $\hat{s}_{t+i+1}$, $\sigma_{t+i+1}$, $\hat{r}_{t+i+1}$ sampled from $p^*_{s(t+i+1)}$ and $a^*_{t+i}$, offers a wealth of predicted information. FUT intends to detect whether after taking $a^*_{t+i}$ and reaching a new state $s_{t+i+1} \neq  \hat{s}_{t+i+1}$ the rest of the predictions in $\tau_{t+i+1}$ still hold.
Thus, we project the trajectory $\tau'_{t+i+1}$ from state \(s_{t+i+1}\) using imagined actions $a^*_{t+i:H} \in \tau_{t+i+1}$ and then we compare $\tau_{t+i+1}$ and $\tau'_{t+i+1}$. If they differ, then the agent is deviated from the plan and it should trigger a replanning event. Otherwise, it proceeds to take action $a_{t+i+1}$.
As long as the new trajectory is similar to the original estimation, we assume that the original plan is still valid and we skip that step. This does not mean that the optimal set of actions at each step is replanned. Rather, every time a replanning is skipped, we propagate only one trajectory starting from the current state of the simulator and still using the originally set of actions, $a^*$, as initially planned.
Algorithm~\ref{alg:fut_filter} describes FUT method.
The original trajectory is estimated with the probability distribution $p^*_{s(t+1)}$ and the updated plan with $p'_{s(t+1)}$. We use Kullback–Leibler divergence (KL) to evaluate the change in two distributions after each step in the simulator. We replan when there is larger difference than $\beta$ (a hyper-parameter). We control how far ahead the two distributions are compared by introducing a hyper-parameter: LA (look ahead steps).

\begin{algorithm}
	\caption{FUT: deciding whether to skip replanning}
	\label{alg:fut_filter}
	\small
	\textbf{Input}: $i$, $p^*_{s(t+i+1:H)}$, $\beta$ and $LA$ 

	\begin{algorithmic}[1]
		\STATE Get $p'_{s(t+i+1:H)}$ from $ComputeTrajectoryProbs$ ($s_{t+i}, \hat{f}, a^*_{t+i:H}$)
		\STATE L = min(H, LA - j)	
		
		\STATE $dist\_error = KL(p'_{s(t+i+1:L)} || p^*_{s(t+i+1:L)})$

        \STATE return TRUE \textbf{if} {$(dist\_error < \beta)$} \textbf{else} FALSE
		
	\end{algorithmic}
\end{algorithm}

\textbf{Bound Imagined Cost Horizon Omission (BICHO):}
Figure~\ref{fig:argumentation} (right) shows that when the system reaches equilibrium, the reward stabilizes as well. 
BICHO assumes the expected reward is stable and attempts to determine whether deviations arise after each step in the trajectory.

At each replanning step, we obtain the distribution of rewards $p^*_{r}$ for H steps in the future. Moreover, at each step of the environment, regardless whether the replanning was skipped or not, we project a new trajectory $p'_{r}$ of H steps, starting from state $s_{t}$ which is given by the environment and using actions $a^*_{t+i}$ in the imagined trajectory obtained at the replanning step.
We then compare this two distributions for LA steps ($<$ $H$), which is a hyper parameter to tune. 
Essentially, replanning steps should be skipped as long as the projected reward in the future does not change significantly with respect to the reward expected from the plan $p^*_{r}$. Note that when doing the comparison immediately after the replanning, both trajectories will start from the same state $s_t$. While, as steps are taken without replanning, the imagined $p^*_{r(t+i)}$ reward starts from one imagined state $p^*_{s(t+i)}$ and the projected trajectory starts at the environment state $s_{t+i}$.
We expect this method to work better (in terms of replanning skipped) in environments where the cost has local equilibrium regions.
The requirement for large overlapping regions between consecutive trajectories is not necessary in our approach, as we consider an overlap of LA steps ahead, which is typically smaller than the trajectory horizon (H).

\begin{algorithm}
	\caption{BICHO: deciding to avoid replanning}
	\label{alg:ccm_filter}
	\small
	\textbf{Input}: $i$, $p^*_{r(t+1:H)}$, $\beta$ and $LA$ 
	\begin{algorithmic}[1]
		\STATE Get $p'_{r(t+i+1:H)}$ from $ComputeTrajectoryProbs$ ($s_{t}, \hat{f}, a^*_{t+i:H}$)
		\STATE L = min(H, LA - i)	
		
		\STATE $r\_error = KL(p'_{r(t+i+1:L)} || p^*_{r(t+1:L)})$

        \STATE return TRUE \textbf{if} {$r\_error < \beta$} \textbf{else} FALSE 	
	\end{algorithmic}
\end{algorithm}

%% file: sections/experiments.tex
\section{Experiments}

\begin{figure*}[ht] 
	\begin{minipage}[b]{0.5\linewidth}
		\centering
		\includegraphics[width=\linewidth]{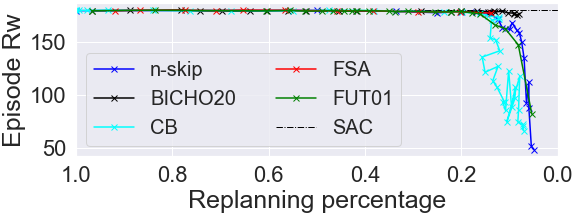}
		\vspace{0ex}
	\end{minipage}
	\begin{minipage}[b]{0.5\linewidth}
		\centering
		\includegraphics[width=\linewidth]{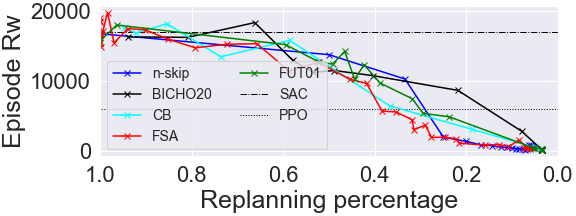}
		\vspace{0ex}
	\end{minipage} 
	\begin{minipage}[b]{0.5\linewidth}
		\centering
		\includegraphics[width=\linewidth]{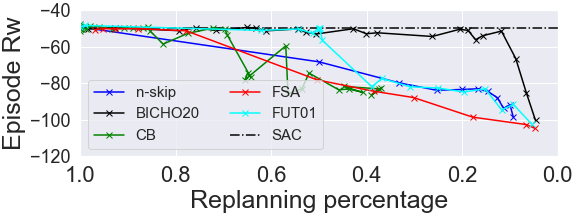}
		\vspace{0ex}
	\end{minipage}
	\begin{minipage}[b]{0.5\linewidth}
		\centering
		\includegraphics[width=\linewidth]{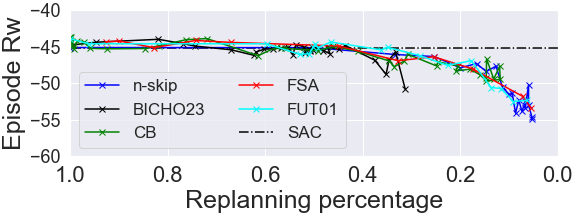} 
		\vspace{0ex}
	\end{minipage} 
	\vspace*{-8mm}
	\caption{Acting upon imagination with pre-trained dynamics. Average of the maximum reward in relation to replanning rate for CP environment using 10 runs for the n-skip, FSA, CB and FUT20 methods. Baseline(n=0), which is PETs, represents 100\% replanning. SAC and PPO at convergence are shown as reference. They are visualized as dotted lines (in x-axis) just for reference.}
	\label{fig:OFF_max}
	\vspace*{-2mm}
\end{figure*}

Our ultimate goal is to reduce computations while controlling performance decay. Intuitively, we expect that a trained dynamics model can anticipate the outcome of the agent's actions and, if its predictions are reliable, it can do so for a number of consecutive imagined steps. So, the first experiment uses pre-trained dynamics to assess the potential gains of acting upon imagined trajectories with the proposed methods: N-Skip (NS), FSA, CB, FUT and BICHO. We aim to investigate the amount of re-planning that can be avoided in terms of number of trajectory planning skipped, the impact of our approach on the reward, and the average and variance of steps executed prior to initiating replanning.

We recognize that acting upon imagination has the potential to afford significant time savings \emph{while training the dynamics models} and can potentially obtain a better result in terms of percentage of re-planning. Therefore, our second experiment evaluates the selected methods online while training the dynamics to assess the savings in overall training time and its effects on performance. In this experiment, we select the best performing methods from the previous experiment to test them while training the dynamics model from scratch. 
We evaluate the methods on agents in the MuJoCo~\cite{conf/iros/TodorovET12} physics engine using two workstations with a last generation GPU. To establish a valid comparison with the baseline PETs ~\cite{kurtl2018deep} (denoted as \textbf{NS1}), we use four environments with corresponding task length ($TaskH$) and trajectory horizon ($H$).\\
We use the following environments: \textbf{Cartpole(CP)}: \(S\in \mathbb{R}^{4}, A \in \mathbb{R}^1\), $TaskH$ 200, $H$ 25. \textbf{Reacher} (RE): \( S \in \mathbb{R}^{17}, A \in \mathbb{R}^7\), $TaskH$ 150, $H$ 25. \textbf{Pusher (PU)}: \( S \in \mathbb{R}^{20}, A \in \mathbb{R}^7\), $TaskH$ 150, $H$ 25. \textbf{HalfCheetah (HC)}: \(S \in \mathbb{R}^{18}, A \in \mathbb{R}^6\), $TaskH$ 1000, $H$ 30.
This means that each iteration will run for $TaskH$, task horizon, steps, and that imagined trajectories include $H$ trajectory horizon steps.
\(S\in \mathbb{R}^{i}, A \in \mathbb{R}^j\) refers to the dimensions of the environment state consisting in a vector of $i$ components and the action consisting in a vector of $j$ components. We assess performance in terms of reward per episode and evaluate wall time and avoided re-planing. All experiments use random seeds and randomize initial conditions per task.

\subsection{Experiment I: Pre-trained Dynamics}
This experiment uses a trained model and compares the uncertainty estimation methods: NS, FSA, CB, FUT, BICHO. It is expected that a pre-trained dynamics model predicts reliably outcomes of immediate actions and can project a number of steps $i<H$ reliably, and any variability in $i$ is attributable to task complexity. We quantify $i$ for each method and environment (see appendix) and also the corresponding percentage of replanning.

A dynamics model is pre-trained for each environment by running Algorithm \ref{alg:algorithm} (no skip) from scratch five times and selecting the best performing model. As a result, we obtain one dynamics model, parameters and replay buffer per task, which we use to evaluate our methods.
For each method, the different hyperparameters are empirically evaluated to find out how robust are the algorithms with respect to hyper-parameters across the different environments.
We report the amount of replanning and the corresponding performance in terms of reward per episode.

We validate each method hyper-parameter with 10 runs per task with different random seeds and randomized initial conditions. We report the episode reward as the maximum reward obtained by the agent in an episode over 10 runs. For NS, we used $n \in \{0-19\}$ steps, where $n = 0$ recalculates at every step and it will be used as a baseline comparison. For FSA, we use constants $c=\{0.1, 0.15, 0.25, 0.35, 0.5, .75, 0.85, 0.9, .95,  .99, .999\}$. The error distribution $\mathbb{E}$ is constructed from error of the state prediction and the actual state in the environment, using the data set obtained while pre-training the dynamics model. 
As the collected errors did not follow a normal distribution, the percentiles will be used in the Algorithm~\ref{alg:fac_filter} to determine whether a trajectory should be recalculated.
In \emph{CB}, we evaluate different values of $C_{cb}=\{0.0,...,2.00\}$ in steps of $0.05$. We expect that a higher value of $C_{cb}$ will make the algorithm decrease the performance, on the other hand a very low value of $C_{cb}$ will make the the algorithm too selective and will result in no skipping at all.
In \emph{FUT} and \emph{BICHO}, we evaluate different values of $\beta=\{0.025, ..., 1024\}$ for different value of look ahead steps, $C_{FUT|BICHO}$, ranging from 1 to $H$. We report the best $c_{FUT|BICHO}$ for the full range of $\beta$.

\subsubsection{Results.} For comparison we added the results for Soft-actor-critic (SAC) ~\cite{haarnoja2018soft} and Proximal Policy Optimization (PPO) ~\cite{schulman2017proximal} at convergence. See Appendix Table 1 
for detailed results for each environment.\\
\textbf{CP.} Fig \ref{fig:OFF_max} (left) shows the performance for \emph{NS}, \emph{FSA}, \emph{CB}, \emph{FUT} and \emph{BICHO} in CP. There is no visible performance degradation by replanning only 40\% of the time, and from 40\% to 20\% the task is still solved with a minor hit in performance. From 20\% the results start decreasing dramatically. Interestingly, in \emph{BICHO} there is no drastic loss in performance even when only replanning 9\% of the steps. This is very close to the limit of the trajectory horizon $H$. \\
\textbf{HC.} Fig \ref{fig:OFF_max} (mid-left) shows the performance in HC, a more complex environment. The graph shows no impact on performance for \emph{FSA}, \emph{CB}, \emph{FUT} and \emph{BICHO} when replanning up to 80\% whilst performance is still acceptable better than SAC at convergence. With less than 60\% replanning, the performance drops drastically. However, \emph{BICHO} still outperforms PPO when recalculating only 20\%.
\emph{N-skip} cannot reduce more than $50\%$ (n=1) without drastically degrading performance, showing that an adaptive method is necessary to skip replanning in complex environments such as HC.\\
\textbf{PU.} Fig \ref{fig:OFF_max} (mid-right) shows the PU results revealing that \emph{FSA}, \emph{CB} and \emph{n-skip} have a stable performance with 80\% replanning and a drastic drop when replanning decreases further. \emph{FUT} keeps a good performance up to 50\% and then the performance starts decreasing drastically. \emph{BICHO} slightly drops in performance after 40\% but it still maintains a good performance by replanning only 10\% of the steps.\\
\textbf{RE.} Fig \ref{fig:OFF_max} (right) shows the results in RE. It reveals no visible performance degradation by replanning only 40\% of the time and from 45\% to 30\% the task is still solved with \emph{FSA} and \emph{CB} with a minor hit in performance. With less than 20\%, reward decreases dramatically. The performance of \emph{FUT} drops after 30\%. In this environment, \emph{n-skip} has comparable degradation with other adaptive methods. However, the fact that \emph{n-skip} starts at $50\%$ and it is fixed should be considered. The adaptive methods working at around $40\%$ replanning still retain performances above SAC and PPO.\\
In all environments, methods projecting future actions achieve longer sequences of steps without replanning with acceptable loss in reward. Results show that replanning less than $70\%$ is feasible whilst retaining state-of-the-art performance. In environments with lower dimensional actions, state spaces, or lower complexity, it is possible to save up to $80\%$ replanning steps. While blindly skipping replanning has an effect, the adaptive methods offer a reasonable trade-off to tune them to work at levels of replanning and performance not reachable by \emph{n-skip}.

\begin{figure*}[hbt]
	\begin{minipage}[b]{0.5\linewidth}
		\centering
		\includegraphics[width=\linewidth]{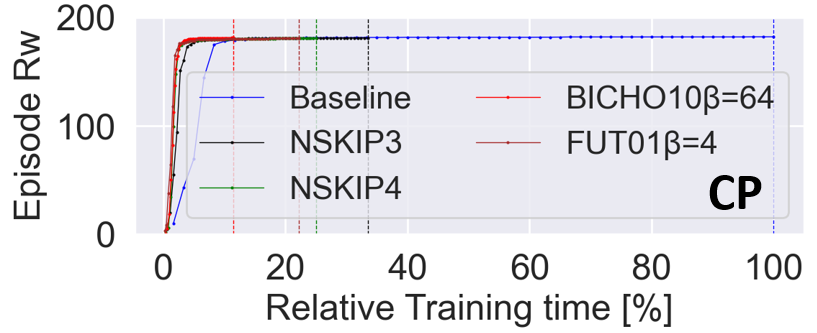}
		\vspace{0ex}
	\end{minipage}
	\begin{minipage}[b]{0.5\linewidth}
		\centering
		\includegraphics[width=\linewidth]{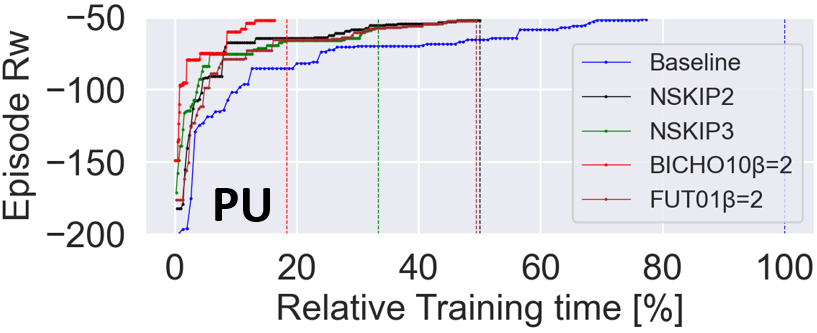}
		\vspace{0ex}
	\end{minipage}
	\begin{minipage}[b]{0.5\linewidth}
		\centering
		\includegraphics[width=\linewidth]{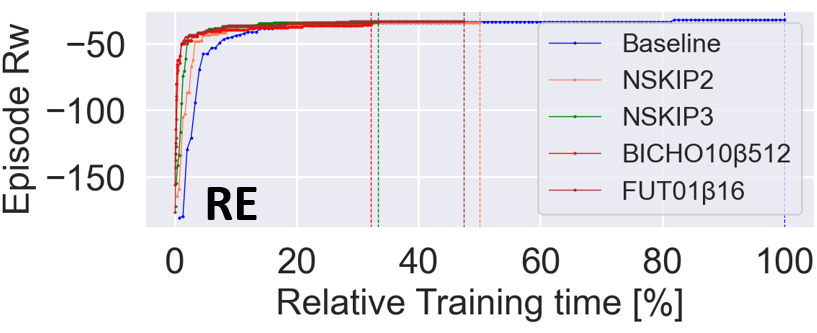}
		\vspace{0ex}
	\end{minipage}
	\begin{minipage}[b]{0.5\linewidth}
		\centering
		\includegraphics[width=\linewidth]{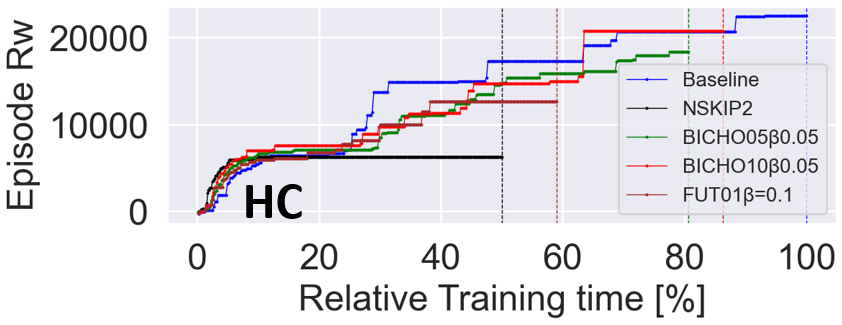}
		\vspace{0ex}
	\end{minipage}
	\vspace*{-9mm}
	\caption{Performance while training the dynamics model. Episode reward in relation to the number of the relative wall time. The vertical lines represent the end of training for the given method. Baseline(n=0) is PETs and represents 100\% recalculation.
	}
	\vspace*{-2mm}
\label{fig:ON_cum_max_reward_relative}

\end{figure*}

\subsection{Experiment II: Online dynamics update}
As the model is being trained, the number of outcomes it can predict reliably varies. Here, uncertainty estimation should result in aborting  plans in favor of re-planning at early training whilst executing longer trajectories as training progresses.

We evaluate our methods while training the dynamics model, using Algorithm \ref{alg:aui_algorithm} with \emph{FUT} and \emph{BICHO}. These methods were selected due to the performance in Experiment I, and because they do not need an error model trained in advance. So we can approximate a real-deployment of the method with minimum tuning effort. 
We evaluate the algorithm in each environment with 3 runs. We increased 50\% the number of training iterations (episodes) in order to better observe the effects of skipping re-planning.
PETs from \cite{kurtl2018deep} is the baseline. 

\subsubsection{Results} In the following, we outline the performance of used algorithms with the best performing hyper-parameters:\\
\textbf{CP.} Fig~\ref{fig:ON_cum_max_reward_relative} shows the results of \emph{BICHO} and \emph{FUT} versus the baseline and NSKIP3 and NSKIP4 depicting the relative wall time compared with the non skip. As \emph{NSKIP3} replans 33\% of the steps, it takes approximate a third of the time to train compared with the baseline, neglecting the training times of the dynamics model. Both \emph{BICHO} and \emph{FUT} outperform the baseline and static skip methods. They retain the same performance of NSKIP by replanning only $15\%$ of the steps. 
In \emph{BICHO}, we observe a negligible impact in performance while reducing the wall time to $10\%$ and reaching top performance within few episodes. To achieve this performance, the conventional MPC needs $200$ calls to $calculateTrajectory$. Instead, \emph{BICHO} performed 28 (SD=1.68). See Table 6
in the appendix. As the dynamics model improves, it produces more accurate predictions and the percentage of replanning steps drops. Indeed, after 10 episodes the replanning needed drop to $10\%$.\\
\textbf{PU.} \emph{BICHO} outperforms the baseline, static methods and \emph{FUT}. It maintains the performance of no skip whilst replanning only $10\%$ of the steps and reducing the wall time to $17\%$, reaching peek performance in few episodes (see Fig~\ref{fig:ON_cum_max_reward_relative}). \\
\textbf{RE.} Fig~\ref{fig:ON_cum_max_reward_relative} shows the results of the \emph{BICHO} and \emph{FUT} methods versus the baseline and NSKIP2 and NSKIP3. \emph{FUT} outperforms the baseline and static skip methods whereas \emph{BICHO} is more conservative in this environment and needs more replanning to reach top performance. Both \emph{BICHO} and \emph{FUT} maintain the performance of no skip, but \emph{BICHO} needs 48\% replanning events whereas \emph{FUT} needs only 14\%.\\
\textbf{HC.} Fig~\ref{fig:ON_cum_max_reward_relative} shows the results of \emph{BICHO} and \emph{FUT} versus the baseline and NSKIP2. Both \emph{BICHO} and \emph{FUT} outperform the static skip methods. \emph{BICHO} has slightly worse top performance than the non skip baseline but it skips 20\% of the replanning steps. \emph{FUT} can skip more steps but the reward drops sharply. In this environment, each method reaches a local optima before it continues improving. Our methods reach this point very quickly without a performance drop and skipping up to 40\% (\textit{FUT}) of the replanning steps.\\
The graphs clearly illustrate significant savings in training times when acting upon imagination. In some cases, these savings are achieved without loss in reward and minimal loss in other cases. More importantly, this loss is referred to training running five or six times longer in the baseline.

%% file: sections/discussion.tex
\section{Discussion}
Acting upon imagination advocates for trusting a reliable imagined trajectory for several steps.
Our experiments show that it leads to 20\%-80\% reduction in computation, depending on the environment, maintaining acceptable reward. The proposed methods leverage different kinds of information available after taking an action in the environment:
\textit{FSA} and \textit{CB} decide to act on the basis of evaluating the last action in a trajectory, \textit{FUT} and \textit{BICHO} evaluate planned future actions from the new state. The latter result in less replanning.
The proposed methods apply to a range of dynamics models for reducing computation costs, regardless of their capabilities to output the uncertainty. FUT and BICHO can be used along any dynamics model that models uncertainty. On the other hand, FSA and CB could be used along any MBRL algorithm by computing statistics of a sliding window of past experiences.
The choice of which algorithm to use depends on the dynamics model's nature. If the dynamics model does not provide a notion of uncertainty, then FSA would be preferable. Otherwise, the methods looking towards the future (FUT, BICHO) have superior performance in terms of saving computation and stability in performance (solving the problem). BICHO reduces most calculations of these last methods while performing at least as the baseline.

\paragraph{Planning}
Skipping replanning can indeed be particularly beneficial in robotics, where hardware limitations often impose constraints on computational resources. By intelligently skipping unnecessary replanning events, we can allocate computational power more efficiently and potentially leverage more sophisticated models or algorithms.

\paragraph{Uncertainty estimation} One alternative avenue of research basing on the work of ~\cite{Zhu2020} would be a progressive measure of mutual information between imagined and real trajectory updated with each successive step to decide whether or not to re-plan. As a limitation, the error expressed as euclidean distance between two state vectors is simple and useful, but it may give misleading information. Comparably, the method BICHO that looks for deviations between imagined reward and a more actual reprojection of future rewards achieves superior performance. Perhaps, a more sophisticated method can be used, taking as input the two state vectors, actions, predicted and observed reward, to output a decision to act or re-plan. For example, a model-free trained policy.

The proposed methods have important implications beyond a greedy motivation to reduce computational effort and time complexity. These methods offer a way to assess how well the dynamics model is at predicting the outcomes of the agent's actions. Conversely, the proposed methods offer a way to evaluate experiences. Meaning, experiences where the outcome of the environment deviates from the predictions of the model may be more informative towards training.
Indeed, our work offers interesting insights on using our method for guided exploration. Assuming steps in an imagined trajectory can be trusted, their evaluation yields a small error, meaning the dynamics model successfully predicts these transitions. One could refine the exploration by omitting actions that lead to transitions with low error, and thus favour less known transitions for future training.
We assume that re-planing is due to errors in the dynamics model. Each re-planning adjusts for the new $H$ with a receding horizon task. But as we act upon imagination, the horizon is no longer receding, and the trajectory risks becoming obsolete. So whence should the model adjust towards a new horizon of imagination?

\section{Conclusion}

In conclusion, our study provides a comprehensive analysis and discussion quantifying the error of predicted trajectories in MBRL. We propose methods for online uncertainty estimation in MBRL, incorporating techniques that observe the outcome of the last action in the trajectory. These methods include comparing the error after performing the last action with the standard expected error and assessing the deviation with respect to expected outcomes using model uncertainty. Additionally, we introduce methods that exploit the forward propagation of the dynamics model to evaluate if the remainder of the plan aligns with expected results and assess the remainder of the plan in terms of the expected reward. These methods update the uncertainty estimation in real-time to assess the utility of the plan.
We demonstrate the efficacy of these uncertainty estimation techniques. Our methods not only leverage accurate predictions but also intelligently determine when to replan trajectories. This approach significantly reduces training time and optimizes the utilization of computational resources by eliminating unnecessary replanning steps. Overall, our findings highlight the potential of these methods to enhance the performance and efficiency of sampling-based MBRL approaches.

%% file: sections/appendices.tex
\section{Environments}
We evaluate the methods on agents in the MuJoCo~\cite{conf/iros/TodorovET12} physics engine. To establish a valid comparison with ~\cite{kurtl2018deep} we use four environments with corresponding task length ($TaskH$) and trajectory horizon ($H$). 
\begin{itemize}
\item Cartpole (CP): \(S\in \mathbb{R}^{4}, A \in \mathbb{R}^1\), $TaskH$ 200, $H$ 25
\item Reacher (RE): \( S \in \mathbb{R}^{17}, A \in \mathbb{R}^7\), $TaskH$ 150, $H$ 25
\item Pusher (PU): \( S \in \mathbb{R}^{20}, A \in \mathbb{R}^7\), $TaskH$ 150, $H$ 25
\item HalfCheetah (HC): \(S \in \mathbb{R}^{18}, A \in \mathbb{R}^6\), $TaskH$ 1000, $H$ 30
\end{itemize}

This means that each iteration will run for $TaskH$, task horizon, steps, and that imagined trajectories include $H$ trajectory horizon steps.
\(S\in \mathbb{R}^{i}, A \in \mathbb{R}^j\) refers to the dimensions of the environment state consisting in a vector of $i$ components and the action consisting in a vector of $j$ components.

\section{Trajectory Quality Analysis}
\begin{figure}[H]
\centering
	\includegraphics[width=0.6\columnwidth]{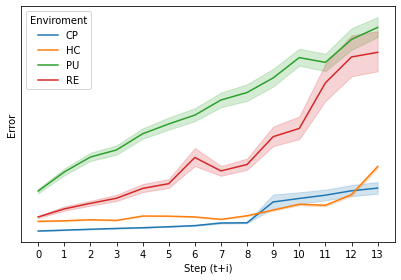}	
	\caption[]{Trajectories error as a function of predicted future steps for all environments.}
	\label{fig:erros_all}    
\end{figure}

The error (euclidean distance between the actual state and predicted states) as a function of the predicted steps in the future is given in Figure~\ref{fig:erros_all}. This Figure is an extension of Figure 2 
to all the environments. One curious observation is that the error for environments PU and RE his relatively higher also when not skipping replanning and increases faster than for environments CP and HC. 

\clearpage

\section{Computational Costs}
\label{axp:computational_costs}
Our proposed algorithms aim to save computations by omitting trajectory recalculations. The complexity of trajectory recalculation is O(H x A x K), H is the length of horizon (we use H=20), A the dimensions of action ($A_{CP}$=1, $A_{RE}$=7, $A_{PU}$=7, $A_{HC}$=6) and K is the number of trajectories generated at each recalculation. K depends on the solver and environment, and in our case it is $K_{CP}$=10000, $K_{HC}$=12500, $K_{PU}$=12500, $K_{RE}$=10000. 

However, the algorithms for deciding whether to skip replanning introduce additional computations. For the n-skip, the computational costs are O(0). Both FSA and CB have a computational complexity $\mathcal{O}(S)$ (computing the error is $\mathcal{O}(S)$ and deciding whether to skip is $\mathcal{O}(0)$), where $S$ is the number of dimensions of the state and it is different for each environment ($S_{CP}$=4, $S_{RE}$=17, $S_{PU}$=20, $S_{HC}$=18). FUT and BICHO project one additional trajectory of length $t \leq H$ to decide whether to skip or not, where $t$ is a hyperparameter. The resulting computational complexity is then $\mathcal{O}(t)$. Additionally, comparing the trajectories increases the complexity by $\mathcal{O}(t)$.
Comparing the computation costs above, it is clear that the costs introduced by n-skip, FCA and CB are negligible compared to the costs of replanning. The cost of both FUT and BICHO is higher but still not comparable to the computational cost of having to replan.

\section{EX-1: Offline Reward and Replanning Rate}
Additional information for 10 runs of each environment for each considered hyper parameter is provided as reference. Table \ref{tbl:res_skip_small} summarizes selected results. Table \ref{tbl:res_skip_CP}, Table \ref{tbl:res_skip_HC}, Table \ref{tbl:res_skip_PU} and Table \ref{tbl:res_skip_RE} show detailed results of each hyper parameter for the environments CP, HC, PU and RE respectively. We report the average and STD reward per episode, steps replanned and number of sequential steps skipped. We also included the error and its STD.

\clearpage

\begin{table}[H]
\centering
\begin{tabular}{@{}lrrrrr@{}}
\toprule
\multicolumn{5}{c}{CP}\\ \midrule
Method  & \multicolumn{1}{c}{RwMax} & \multicolumn{1}{c}{Rw}  & \multicolumn{1}{c}{Rc} & \multicolumn{1}{c}{RcPer} & \multicolumn{1}{c}{i}\\
\midrule
Baseline & 179.373 & 178.830  & 200.00  &   1.00 &   0.00  \\
NSKIP1   & 179.610 & 178.201  & 100.00  &   0.50 &   0.99   \\
NSKIP2   & 178.687 & 177.474  &  67.00  &   0.33 &   1.97   \\
FSA0.50  & 179.923 & 178.997  & 130.50  &   0.65 &   1.25  \\
FSA0.99  & 177.061 & 172.569  &  27.60  &   0.13 &   6.22   \\
CB0.50   & 179.473 & 178.303  &  76.50  &   0.38 &   2.77  \\
BICHO10$\beta$32 & 179.061 & 175.951 &  21.70  & 0.10  &   7.92   \\
BICHO20$\beta$64 & 177.022 & 174.542  &  18.00 & 0.09 &   9.79  \\
FUT01$\beta$4.00  & 178.620 & 175.474 &  40.000 &   0.20 &   3.94 \\
\midrule

\multicolumn{5}{c}{HC} 	\\
\midrule
Baseline        & 16750.776   &     12764.668 &   1000.00  &   1.000  &    0.000 \\
NSKIP1          & 13748.625   &   9247.623    &    500.00  &   0.500  &    0.998 \\
NSKIP2          & 10266.311   &   6118.676    &    334.00  &   0.334  &    1.994 \\
FSA0.50         & 15791.490   &   10881.401   &    586.00      &   0.586  &    1.613 \\
BICHO05$\beta$0.200   & 18341.042   &   11637.706   &   662.30     &   0.662  &    0.998 \\
BICHO05$\beta$6.000   & 8646.361    &   1872.543    &   218.50     &   0.218  &    7.004 \\
FUT01$\beta$0.100     & 15190.099   &   10127.721   &   595.10     &   0.595  &    1.021 \\
\midrule

\multicolumn{5}{c}{PU} 	\\	 
\midrule
Baseline            &   -49.277 & -56.858 & 150.00  & 1.000 &  0.000 \\
NSKIP1              &   -68.296 & -79.493 & 75.00   & 0.500 &  0.990 \\
NSKIP2              &   -79.710 & -85.368 & 50.00   & 0.333 &  1.960 \\
FSA0.50             &   -51.347 & -78.970 & 116.70  & 0.773 &  1.199 \\
CB1.00              &   -49.149 & -76.829 & 128.70  & 0.853 &  1.242 \\
BICHO10$\beta$16.0  &   -51.527 & -85.723 & 17.70   & 0.118 &  7.002 \\
FUT01$\beta$0.40    &   -56.583 & -81.134 & 73.90   & 0.493 &  1.019 \\
\midrule
 
\multicolumn{5}{c}{RE} 	\\	 
\midrule
Baseline        & -45.121 & -45.930  & 150.00   & 1.000  & 0.000 \\
NSKIP1          & -45.144 & -46.296 & 75.00    & 0.500  & 0.987 \\
NSKIP2          & -46.076 & -47.167 & 50.00    & 0.333 & 1.960 \\
FSA0.50         & -44.420  & -46.080  & 100.60 & 0.671 & 1.211 \\
CB1.00          & -45.097 & -46.592 & 86.50  & 0.577 & 2.011 \\
BICHO23$\beta$0.125   & -44.609 & -45.972 & 115.10 & 0.767 & 0.972 \\
BICHO23$\beta$768     & -45.719 & -58.677 & 49.90  & 0.333 & 2.154 \\
FUT01$\beta$0.200     & -45.080  & -46.554 & 75.20  & 0.501 & 0.991 \\
FUT01$\beta$8.000     & -46.922 & -49.804 & 26.60  & 0.177 & 4.549 \\
\midrule
\end{tabular}
\caption{Pre-Trained Dynamics model. Performance of selected models. Rw is average reward. Rc is the number of replanning events, RcPer is the percentage of replanning, i the average number of consecutive steps executed before replanning. }
\label{tbl:res_skip_small}
\end{table}

\begin{table}[hb]
	\small
	\begin{tabular}{@{}llllllllll@{}}
		\toprule
		\multicolumn{10}{c}{CP}                                                                                                                                                                                                                                                               \\ \midrule
		Method  & \multicolumn{1}{c}{Rw} & \multicolumn{1}{c}{RwSTD} & \multicolumn{1}{c}{Rc} & \multicolumn{1}{c}{RcSTD} & \multicolumn{1}{c}{RcPer} & \multicolumn{1}{c}{i mean} & \multicolumn{1}{c}{i STD} & \multicolumn{1}{c}{RwMin} & \multicolumn{1}{c}{RwMax} \\ \midrule
		Baseline & 178.83 & 0.41 & 200.00 & 0.00 & 1.00 & 0.00 & 0.00 & 177.92 & 179.37 \\
		NSKIP1 & 178.20 & 0.84 & 100.00 & 0.00 & 0.50 & 0.99 & 0.00 & 176.62 & 179.61 \\
		NSKIP2 & 177.47 & 0.90 & 67.00 & 0.00 & 0.34 & 1.97 & 0.00 & 175.87 & 178.69 \\
		NSKIP3 & 175.91 & 2.72 & 50.00 & 0.00 & 0.25 & 2.94 & 0.00 & 168.24 & 177.43 \\
		NSKIP5 & 174.26 & 4.74 & 34.00 & 0.00 & 0.17 & 4.85 & 0.00 & 161.63 & 177.95  \\
		NSKIP6 & 170.22 & 6.35 & 29.00 & 0.00 & 0.14 & 5.79 & 0.00 & 162.63 & 176.79 \\
		NSKIP7 & 163.60 & 4.70 & 25.00 & 0.00 & 0.12 & 6.72 & 0.00 & 155.70 & 171.89  \\
		NSKIP8 & 158.94 & 5.97 & 23.00 & 0.00 & 0.12 & 7.65 & 0.00 & 143.23 & 163.29  \\
		NSKIP9 & 129.26 & 35.15 & 20.00 & 0.00 & 0.10 & 8.55 & 0.00 & 68.02 & 161.79 \\

		BICHO10$\beta$0.10 & 178.76 & 0.75 & 177.60 & 5.25 & 0.89 & 0.95 & 0.03 & 177.12 & 179.69  \\
		BICHO10$\beta$0.20 & 178.81 & 0.55 & 156.30 & 4.64 & 0.78 & 0.98 & 0.01 & 178.05 & 179.98  \\
		BICHO10$\beta$0.40 & 178.29 & 0.46 & 148.70 & 8.71 & 0.74 & 1.01 & 0.21 & 177.53 & 178.84  \\
		BICHO10$\beta$0.80 & 178.05 & 0.64 & 139.60 & 3.89 & 0.70 & 1.02 & 0.31 & 176.98 & 178.74  \\
		BICHO10$\beta$32 & 177.71 & 0.89 & 83.90 & 15.16 & 0.42 & 1.70 & 0.84 & 176.48 & 179.06    \\
		BICHO10$\beta$128 & 177.78 & 0.50 & 44.40 & 11.71 & 0.22 & 3.79 & 0.56 & 176.90 & 178.52  \\

		FSA0.15 & 178.93 & 0.53 & 173.60 & 4.81 & 0.87 & 1.02 & 0.07 & 178.03 & 179.74 \\
		FSA0.25 & 178.87 & 0.66 & 154.60 & 3.63 & 0.77 & 1.10 & 0.09 & 177.95 & 179.76  \\
		FSA0.35 & 178.87 & 0.37 & 141.80 & 4.19 & 0.71 & 1.18 & 0.08 & 177.98 & 179.43  \\
		\textbf{FSA0.50} & 179.00 & 0.57 & 130.50 & 4.97 & \textbf{0.65} & 1.25 & 0.08 & 177.92 & \textbf{179.92} \\
		FSA0

.99 & 172.57 & 5.24 & 27.60 & 2.59 & 0.14 & 6.23 & 0.40 & 161.68 & 177.06  \\
		\textbf{CB0.50} & 178.30 & 1.14 & 76.50 & 6.36 & \textbf{0.38} & 2.77 & 0.50 & 175.34 & \textbf{179.47} \\
		CB0.90 & 120.29 & 45.19 & 27.50 & 7.50 & 0.14 & 8.78 & 0.62 & 31.59 & 170.37 \\
		CB1.00 & 114.16 & 34.08 & 25.80 & 5.94 & 0.13 & 9.27 & 0.62 & 44.18 & 151.96 \\
		CB1.75 & 74.17 & 26.98 & 15.80 & 4.59 & 0.08 & 14.43 & 0.48 & 40.05 & 117.84 \\

		BICHO20$\beta$0.05 & 178.76 & 0.75 & 177.60 & 5.25 & 0.89 & 0.95 & 0.03 & 177.12 & 179.69 \\
		BICHO20$\beta$0.10 & 178.81 & 0.55 & 156.30 & 4.64 & 0.78 & 0.98 & 0.01 & 178.05 & 179.98 \\
		BICHO20$\beta$0.70 & 177.80 & 0.61 & 56.90 & 15.60 & 0.29 & 2.87 & 0.87 & 176.95 & 178.81 \\
		BICHO20$\beta$1 & 177.60 & 0.73 & 40.20 & 7.71 & 0.20 & 4.07 & 0.60 & 176.56 & 178.57 \\
		BICHO20$\beta$8 & 177.02 & 0.83 & 26.00 & 0.82 & 0.13 & 6.48 & 0.39 & 175.50 & 178.13 \\
		\textbf{BICHO20$\beta$64} & 172.28 & 7.54 & 18.30 & 1.83 & \textbf{0.09} & 9.70 & 0.58 & 153.22 & \textbf{177.02} \\
		BICHO20$\beta$256 & 165.47 & 14.67 & 17.00 & 1.25 & 0.08 & 10.47 & 0.97 & 138.41 & 175.40 \\

		BICHO23$\beta$0.05 & 178.69 & 0.41 & 178.00 & 3.62 & 0.89 & 0.96 & 0.02 & 178.00 & 179.30 \\
		BICHO23$\beta$0.10 & 178.37 & 0.91 & 150.80 & 4.05 & 0.75 & 0.99 & 0.03 & 176.97 & 179.57 \\
		BICHO23$\beta$0.70 & 177.88 & 0.40 & 48.80 & 17.96 & 0.24 & 3.47 & 0.90 & 177.34 & 178.56 \\
		BICHO23$\beta$1 & 177.46 & 0.69 & 42.50 & 14.32 & 0.21 & 3.99 & 0.86 & 176.23 & 178.42 \\
		BICHO23$\beta$8 & 175.01 & 7.34 & 25.30 & 1.64 & 0.13 & 6.67 & 0.46 & 154.18 & 177.88 \\
		BICHO23$\beta$64 & 174.54 & 2.08 & 18.00 & 0.67 & 0.09 & 9.79 & 0.48 & 170.59 & 177.01 \\

		FUT01$\beta$0.05 & 178.61 & 0.64 & 167.30 & 4.88 & 0.84 & 0.97 & 0.01 & 177.55 & 179.70 \\
		FUT01$\beta$0.15 & 178.45 & 0.54 & 104.80 & 1.48 & 0.52 & 1.04 & 0.05 & 177.44 & 179.04 \\
		FUT01$\beta$0.80 & 177.08 & 0.81 & 67.50 & 1.78 & 0.34 & 1.95 & 0.05 & 175.96 & 178.15 \\
		FUT01$\beta$2.00 & 177.19 & 0.75 & 50.80 & 0.92 & 0.25 & 2.91 & 0.04 & 176.18 & 178.07 \\
		\textbf{FUT01$\beta$4} & 175.47 & 4.03 & 40.00 & 0.47 & \textbf{0.20} & 3.94 & 0.09 &

 164.78 & \textbf{178.62} \\
		FUT01$\beta$64.0 & 82.46 & 34.04 & 16.40 & 1.43 & 0.08 & 10.92 & 0.78 & 27.88 & 147.24 \\
		FUT01$\beta$256  & 23.74 & 24.51 & 10.80 & 1.48 & 0.05 & 17.11 & 0.72 & 4.45 & 82.34 \\
		\bottomrule
	\end{tabular}

	\caption{Performance of each method in the environment CP. Rw is average reward. RC is the number of replanning events, RcPer is the percentage of replanning, i is the depth of trajectory used, the number of consecutive steps executed before replanning.}
	\label{tbl:res_skip_CP}
\end{table}

\clearpage

\begin{table*}[hb]
		\small
    \begin{tabular}{@{}llllllllll@{}}
        \toprule
        \multicolumn{10}{c}{HC}                                                                                                                                                                                                                                                               \\ \midrule
        Method  & \multicolumn{1}{c}{Rw} & \multicolumn{1}{c}{RwSTD} & \multicolumn{1}{c}{Rc} & \multicolumn{1}{c}{RcSTD} & \multicolumn{1}{c}{RcPer} & \multicolumn{1}{c}{i mean} & \multicolumn{1}{c}{i STD} & \multicolumn{1}{c}{RwMin} & \multicolumn{1}{c}{RwMax}  \\ \midrule
Baseline  & 12764.668              & 2849.853                  & 1000.000               & 0.000        & 1.000             & 0.000                  & 0.000                     & 7372                  & 16750                      \\
NSKIP1  & 9247.623               & 2179.981                  & 500.000                & 0.000          & 0.500           & 0.998                  & 0.000                     & 5299                  & 13748                          \\
NSKIP2  & 6118.676               & 3011.780                  & 334.000                & 0.000           & 0.334          & 1.994                  & 0.000                     & 1375                  & 10266                       \\
NSKIP3  & 1443.533               & 357.099                   & 250.000                & 0.000          & 0.250           & 2.988                  & 0.000                     & 1008                  & 1944                     \\
NSKIP4  & 1048.150               & 147.981                   & 200.000                & 0.000         & 0.200            & 3.980                  & 0.000                     & 859                   & 1352                          \\
NSKIP5  & 750.226                & 36.873                    & 167.000                & 0.000        & 0.167             & 4.970                  & 0.000                     & 704                   & 823                       \\
NSKIP7  & 453.360                & 40.811                    & 125.000                & 0.000        & 0.125             & 6.944                  & 0.000                     & 379                   & 512                   \\
NSKIP9  & 261.672                & 146.953                   & 100.000                & 0.000       & 0.100              & 8.910                  & 0.000                     & -11                   & 379                    \\
FSA0.15 & 10637.761              & 3986.620                  & 922.400                & 10.700     & 0.922               & 1.031                  & 0.059                     & 5718                  & 16835                    \\
FSA0.25 & 13283.989              & 4397.851                  & 854.300                & 23.636     & 0.854               & 1.099                  & 0.061                     & 5099                  & 18182                  \\
FSA0.35 & 9963.964               & 2448.046                  & 736.400                & 20.919    & 0.736                & 1.244                  & 0.049                     & 5190                  & 13450                 \\
\textbf{FSA0.50} & 10881.401              & 2565.614                  & 586.000                & 22.691    & \textbf{0.586}               & 1.613                  & 0.078                     & 6997    &  \textbf{15791}        \\
FSA0.90 & 173.839                & 143.729                   & 50.500                 & 11.336    & 0.051                & 19.838                 & 1.536                     & 53                    & 499               \\
CB0.50  & 13269.067              & 3347.259                  & 994.400                & 1.174    & 0.994                & 0.844                  & 0.026                     & 6550                  & 17071               \\
CB0.90  & 11010.053              & 3129.994                  & 657.900                & 8.900    & 0.658                 & 1.612                  & 0.215                     & 4781                  & 15326           \\
\textbf{CB1.00}  & 7262.139               & 3246.655                  & 523.500                & 31.366   & \textbf{0.524}                 & 2.031                  & 0.175                     & 3040  & \textbf{12650}        \\
CB1.75  & 408.850                & 412.640                   & 85.100                 & 37.245   & 0.085                 & 13.553                 & 1.866                     & 105                   & 1538                 \\
BICHO05$\beta$0.050 &12685.246 &3163.449 & 940.100 &   3.414 &   0.940 &   0.984 &   0.004 &5015 &16299 \\
BICHO05$\beta$0.100 &11696.336 &2267.720 & 808.400 &   9.812 &   0.808 &   0.995 &   0.002 &9150 &16227 \\
\textbf{BICHO05$\beta$0.200}   &11637.706 &3778.409 & 662.300 &   8.538 &   \textbf{0.662} &   0.998 &   0.010 &6274 & \textbf{18341} \\
BICHO05$\beta$0.800 &8218.947 &2584.988 & 535.400 &   4.477 &   0.535 &   1.031 &   0.122 &4207 &11222 \\
BICHO05$\beta$4.000 &5302.087 &3623.648 & 404.700 &  40.604 &   0.405 &   1.517 &   1.104 &1293 &10756\\
BICHO05$\beta$6.000 &1872.543 &2639.615 & 218.500 & 130.996 &   0.218 &   7.004 &   3.940 &   5 &8646  \\
BICHO05$\beta$8.000 & 587.539 & 865.745 &  79.200 &  42.856 &   0.079 &  14.786 &   1.851 &  11 &2800\\
FUT01$\beta$0.025 &12936.022 &2014.828 & 999.500 &   0.707 &   1.000 &   0.217 &   0.255 &10037 &15763\\
FUT01$\beta$0.050 &13436.725 &3264.052 & 964.100 &   9.158 &   0.964 &   0.971 &   0.025 &9192 &18000 \\
\textbf{FUT01$\beta$0.100} &10127.721 &2911.831 & 595.100 &  11.396 &   \textbf{0.595} &   1.021 &   0.022 &6630 &\textbf{15190}  \\
FUT01$\beta$0.125 &8075.127 &3342.769 & 522.200 &   5.922 &   0.522 &   1.048 &   0.013 &1733 &12706 \\
FUT01$\beta$0.150 &7652.031 &2645.142 & 492.400 &   7.545 &   0.492 &   1.086 &   0.028 &3945 &12400  \\
FUT01$\beta$0.400 &6498.406 &2534.774 & 388.800 &   4.638 &   0.389 &   1.570 &   0.025 &1645 &9707 \\
FUT01$\beta$2.000 &1973.394 &1416.910 & 237.500 &   7.367 &   0.237 &   3.208 &   0.083 & 878 &4863\\

\bottomrule
\end{tabular}
\caption{Performance of each method in the environment HC. Rw is average reward. RC is the number of replanning events, RcPer is the percentage of replanning, i is the depth of trajectory used, the number of consecutive steps executed before replanning.}
\label{tbl:res_skip_HC}
\end{table*}

\clearpage

\begin{table*}[hb]
		\small
    \begin{tabular}{@{}llllllllll@{}}
        \toprule
        \multicolumn{10}{c}{PU}                                                                                                                                                                                                                                                               \\ \midrule
        Method  & \multicolumn{1}{c}{Rw} & \multicolumn{1}{c}{RwSTD} & \multicolumn{1}{c}{Rc} & \multicolumn{1}{c}{RcSTD} & \multicolumn{1}{c}{RcPer} & \multicolumn{1}{c}{i mean} & \multicolumn{1}{c}{i STD} & \multicolumn{1}{c}{RwMin} & \multicolumn{1}{c}{RwMax}  \\ \midrule
Baseline & -56.858 &  10.768 & 150.000 &   0.000 &   1.000 &   0.000 &   0.000 & -75.277 & -49.277  \\
NSKIP1 & -79.493 &   9.199 &  75.000 &   0.000 &   0.500 &   0.987 &   0.000 & -96.068 & -68.296 \\
NSKIP2 & -85.368 &   3.235 &  50.000 &   0.000 &   0.333 &   1.960 &   0.000 & -89.991 & -79.710 \\
NSKIP3 & -87.983 &   3.633 &  38.000 &   0.000 &   0.253 &   2.921 &   0.000 & -96.177 & -83.838  \\
NSKIP4 & -86.773 &   3.526 &  30.000 &   0.000 &   0.200 &   3.867 &   0.000 & -94.565 & -83.384 \\
NSKIP5 & -89.847 &   5.745 &  25.000 &   0.000 &   0.167 &   4.800 &   0.000 & -96.857 & -83.023\\
NSKIP6 & -89.504 &   3.974 &  22.000 &   0.000 &   0.147 &   5.727 &   0.000 & -94.531 & -84.423 \\
NSKIP7 & -93.925 &   3.331 &  19.000 &   0.000 &   0.127 &   6.632 &   0.000 & -99.409 & -87.872  \\
NSKIP8 & -97.002 &   2.200 &  17.000 &   0.000 &   0.113 &   7.529 &   0.000 &-100.367 & -93.706 \\
NSKIP9 & -97.425 &   3.086 &  15.000 &   0.000 &   0.100 &   8.400 &   0.000 &-101.036 & -91.792 \\
FSA0.15 & -71.232                & 14.667                    & 142.700                & 3.433        &  0.951             & 0.852                  & 0.066                     & -86.917                   & -50.196              \\
FSA0.25 & -71.606                & 12.279                    & 138.900                & 4.748        &  0.920             & 0.916                  & 0.065                     & -86.514                   & -50.309           \\
FSA0.35 & -68.623                & 15.260                    & 131.700                & 10.371       &  0.873             & 0.990                  & 0.057                     & -89.602                   & -50.198             \\
\textbf{FSA0.50} & -78.970                & 16.020                    & 116.700                & 12.859       &  \textbf{0.773}            & 1.199                  & 0.096                     & -100.741                  & \textbf{-51.347}            \\
FSA0.90 & -106.759               & 1.643                     & 7.100                  & 0.316        &  0.047            & 18.775                 & 0.164                     & -109.108                  & -104.638                 \\
CB0.50  & -59.227                & 12.197                    & 149.800                & 0.422        &  0.993             & 0.100                  & 0.211                     & -87.747                   & -50.568                       \\
CB0.90  & -67.436                & 15.815                    & 133.400                & 6.150        &  0.887             & 1.142                  & 0.130                     & -87.785                   & -49.913                \\
\textbf{CB1.00}  & -76.829                & 16.610                    & 128.700                & 8.138        &  \textbf{0.853}             & 1.242                  & 0.212                     & -99.612                   & \textbf{-49.149}               \\
CB1.75  & -90.519                & 6.520                     & 65.600                 & 8.540        &  0.433             & 3.049                  & 0.625                     & -103.549                  & -83.159           \\       
BICHO10$\beta$0.05 & -59.871 &   9.853 & 139.100 &   4.332 &   0.927 &   0.899 &   0.061 & -73.980 & -49.726  \\
BICHO10$\beta$0.35 & -72.783 &  15.044 &  81.700 &   3.889 &   0.545 &   1.079 &   0.467 & -90.490 & -50.358 \\
BICHO10$\beta$0.50 & -75.205 &  10.421 &  75.900 &   8.517 &   0.506 &   1.192 &   0.894 & -89.100 & -53.112 \\
BICHO10$\beta$2.00 & -75.101 &  14.264 &  39.400 &   9.663 &   0.263 &   2.838 &   1.563 & -90.866 & -54.450 \\
BICHO10$\beta$3.00 & -76.374 &  13.973 &  30.800 &   1.317 &   0.205 &   3.471 &   0.422 & -91.605 & -50.523 \\
BICHO10$\beta$8.00 & -78.874 &  13.931 &  23.600 &   2.547 &   0.157 &   5.076 &   0.582 & -93.285 & -54.045 \\
\textbf{BICHO10$\beta$16.0} & -85.723 &  13.360 &  17.700 &   1.636 &   \textbf{0.118} &   7.002 &   0.749 &-100.344 & \textbf{-51.527} \\
BICHO10$\beta$32.0 & -89.359 &  11.471 &  13.200 &   2.394 &   0.088 &   9.805 &   0.474 &-108.031 & -66.529 \\

FUT01$\beta$0.25 & -56.858 &  10.768 & 150.000 &   0.000 &   1.000 &   0.000 &   0.000 & -75.277 & -49.277 \\
FUT01$\beta$0.05 & -54.335 &   6.991 & 148.300 &   1.494 &   0.989 &   0.508 &   0.200 & -68.104 & -48.443 \\
\textbf{FUT01$\beta$0.40} & -81.134 &  10.063 &  73.900 &   0.738 &   \textbf{0.493} &   1.019 &   0.066 & -93.088 & \textbf{-56.583}  \\
FUT01$\beta$0.80 & -86.352 &   3.430 &  58.400 &   2.459 &   0.389 &   1.554 &   0.046 & -92.642 & -81.845 \\
FUT01$\beta$1.00 & -84.547 &   3.920 &  55.200 &   1.476 &   0.368 &   1.703 &   0.048 & -90.966 & -77.338 \\
FUT01$\beta$2.00 & -85.207 &   1.824 &  46.400 &   2.119 &   0.309 &   2.211 &   0.074 & -88.006 & -81.926 \\
FUT01$\beta$4.00 & -88.125 &   4.487 &  37.500 &   1.434 &   0.250 &   2.965 &   0.079 & -96.489 & -82.626 \\
FUT01$\beta$16.0 & -90.654 &   5.113 &  22.800 &   0.919 &   0.152 &   5.484 &   0.116 & -97.510 & -83.038 \\
FUT01$\beta$32.0 & -97.671 &   2.087 &  17.600 &   0.516 &   0.117 &   7.282 &   0.178 &-101.307 & -94.572 \\
FUT01$\beta$64.0 & -99.894 &   4.789 &  14.000 &   0.667 &   0.093 &   9.455 &   0.276 &-104.336 & -91.531 \\
FUT01$\beta$64.0 &-105.005 &   1.319 &   8.200 &   0.422 &   0.055 &  16.189 &   0.301 &-106.247 &-102.273 \\

\\ \bottomrule
\end{tabular}
\caption{Performance of each method in the environment PU. Rw is average reward. RC is the number of replanning events, RcPer is the percentage of replanning, i is the depth of trajectory used, the number of consecutive steps executed before replanning.}
\label{tbl:res_skip_PU}
\end{table*}
\clearpage

\begin{table*}[hb]
		\small
	\begin{tabular}{@{}llllllllll@{}}
		\toprule
		\multicolumn{10}{c}{RE}                                                                                                                                                                                                                                                               \\ \midrule
		Method  & \multicolumn{1}{c}{Rw} & \multicolumn{1}{c}{RwSTD} & \multicolumn{1}{c}{Rc} & \multicolumn{1}{c}{RcSTD} & \multicolumn{1}{c}{RcPer} & \multicolumn{1}{c}{i mean} & \multicolumn{1}{c}{i STD} & \multicolumn{1}{c}{RwMin} & \multicolumn{1}{c}{RwMax}  \\ \midrule
Baseline  & -45.930                & 0.606                     & 150.000                & 0.000      & 1.000               & 0.000                  & 0.000                     & -46.689                   & -45.121                                 \\
NSKIP1  & -46.296                & 0.934                     & 75.000                 & 0.000      &  0.500              & 0.987                  & 0.000                     & -47.843                   & -45.144                                  \\
NSKIP2  & -47.167                & 0.815                     & 50.000                 & 0.000     & 0.333                & 1.960                  & 0.000                     & -48.515                   & -46.076                               \\
NSKIP3  & -48.553                & 1.075                     & 38.000                 & 0.000      & 0.253               & 2.921                  & 0.000                     & -49.748                   & -46.352                                       \\
NSKIP4  & -49.855                & 0.953                     & 30.000                 & 0.000      & 0.200               & 3.867                  & 0.000                     & -51.185                   & -47.895                                    \\
NSKIP5  & -50.072                & 1.763                     & 25.000                 & 0.000     & 0.167                & 4.800                  & 0.000                     & -53.788                   & -47.317                                     \\
NSKIP7  & -51.640                & 2.480                     & 19.000                 & 0.000     & 0.127                & 6.632                  & 0.000                     & -54.880                   & -47.674                                    \\
NSKIP9  & -54.798                & 1.955                     & 15.000                 & 0.000    & 0.100                 & 8.400                  & 0.000                     & -57.343                   & -51.526                                     \\
FSA0.15 & -45.591                & 1.155                     & 134.800                & 2.860    & 0.899                 & 0.958                  & 0.058                     & -48.215                   & -44.236                                       \\
FSA0.25 & -45.639                & 0.512                     & 124.200                & 4.984    & 0.828                 & 1.006                  & 0.071                     & -46.553                   & -45.137                                      \\
FSA0.35 & -45.839                & 1.132                     & 111.400                & 2.875   & 0.743                  & 1.094                  & 0.109                     & -47.770                   & -44.163                                      \\
FSA0.50 & -46.080                & 1.044                     & 100.600                & 2.319   & 0.671                  & 1.211                  & 0.117                     & -47.473                   & -44.420                                     \\
FSA0.90 & -57.192                & 2.688                     & 8.200                  & 0.422   & 0.055                  & 16.254                 & 0.583                     & -63.307                   & -53.410                                     \\
\textbf{CB0.50}  & -46.355                & 0.555                     & 148.700                & 1.337  & \textbf{0.991}                   & 0.422                  & 0.230                     & -47.001                   & \textbf{-45.323}                \\
CB0.90  & -45.709                & 1.114                     & 107.900                & 3.635   & 0.719                  & 1.465                  & 0.449                     & -47.032                   & -43.904                                      \\
\textbf{CB1.00}  & -46.592                & 1.073                     & 86.500                 & 8.303   & \textbf{0.577}                  & 2.011                  & 0.582                     & -48.847                   & \textbf{-45.097}              \\
CB1.75  & -51.375                & 1.911                     & 22.000                 & 3.859  & 0.147    & 8.066                  & 1.230                     & -55.623                   & -49.221                  \\   
BICHO23$\beta$0.100 & -45.586 &   1.053 & 122.800 &   4.517 &   0.819 &   0.964 &   0.013 & -47.478 & -44.001 \\
\textbf{BICHO23$\beta$0.125} & -45.972 &   0.939 & 115.100 &   4.725 &   \textbf{0.767} &   0.972 &   0.011 & -47.803 & \textbf{-44.609} \\
BICHO23$\beta$256   & -47.043 &   0.898 &  72.100 &   1.912 &   0.481 &   1.072 &   0.360 & -48.583 & -45.539  \\
BICHO23$\beta$576   & -53.622 &   4.640 &  56.300 &   8.795 &   0.375 &   1.718 &   1.546 & -62.853 & -46.846  \\
\textbf{BICHO23$\beta$768}   & -58.677 &   7.794 &  49.900 &  12.142 &   \textbf{0.333} &   2.154 &   1.811 & -68.808 & \textbf{-45.719} \\
BICHO23$\beta$1024 & -58.765 &   7.331 &  46.900 &   8.034 &   0.313 &   2.272 &   1.153 & -74.104 & -50.834 \\

FUT01$\beta$0.025 & -46.003 &   1.062 & 150.000 &   0.000 &   1.000 &   0.000 &   0.000 & -47.128 & -43.905 \\
FUT01$\beta$0.050 & -45.394 &   0.577 & 144.300 &   2.541 &   0.962 &   0.822 &   0.061 & -46.137 & -44.614 \\
\textbf{FUT01$\beta$0.200} & -46.554 &   1.034 &  75.200 &   0.422 &   \textbf{0.501} &   0.991 &   0.024 & -48.400 & \textbf{-45.080}  \\
FUT01$\beta$1.000 & -47.508 &   1.035 &  52.300 &   1.418 &   0.349 &   1.851 &   0.058 & -48.661 & -45.099 \\
\textbf{FUT01$\beta$8.000} & -49.804 &   1.872 &  26.600 &   1.075 &   \textbf{0.177} &   4.549 &   0.131 & -53.130 & \textbf{-46.922} \\
FUT01$\beta$32.00 & -53.242 &   1.743 &  17.400 &   0.699 &   0.116 &   7.325 &   0.346 & -55.674 & -50.721  \\
FUT01$\beta$64.00 & -54.556 &   1.604 &  13.700 &   0.823 &   0.091 &   9.669 &   0.420 & -57.550 & -52.503  \\
FUT01$\beta$256.0 & -56.823 &   3.401 &   9.400 &   0.516 &   0.063 &  14.499 &   0.485 & -63.382 & -52.324  \\

\\ \bottomrule
\end{tabular}
\caption{Performance of each method in the environment RE. Rw is average reward. RC is the number of replanning events, RcPer is the percentage of replanning, i is the depth of trajectory used, the number of consecutive steps executed before replanning.}
\label{tbl:res_skip_RE}
\end{table*}
\clearpage

\section{EX-2: Online dynamics update}
\begin{figure}
\centering
\begin{minipage}{0.45\textwidth}
  \centering
  \includegraphics[width=\linewidth]{images/CP_ON_cum_max_reward_relative.png}
  \includegraphics[width=\linewidth]{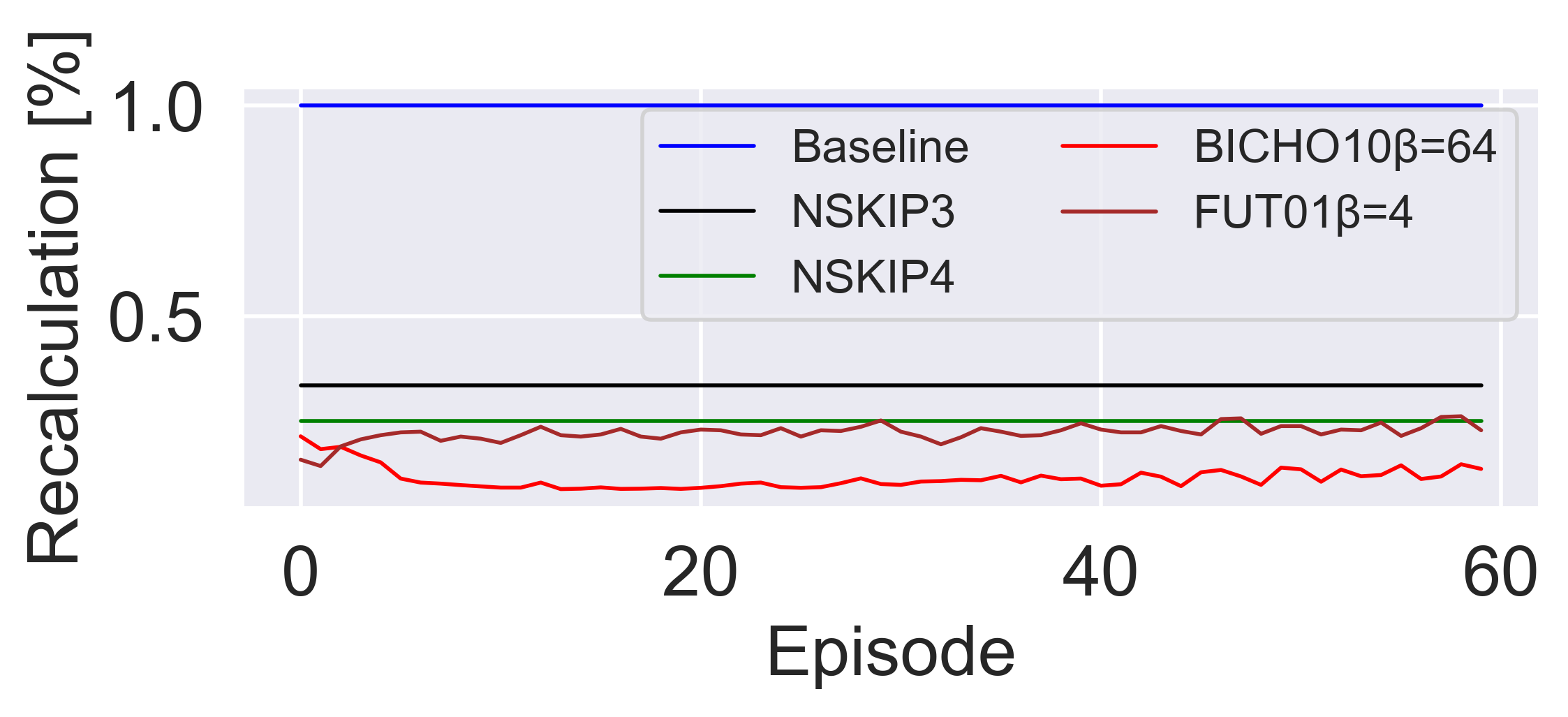}
  \includegraphics[width=\linewidth]{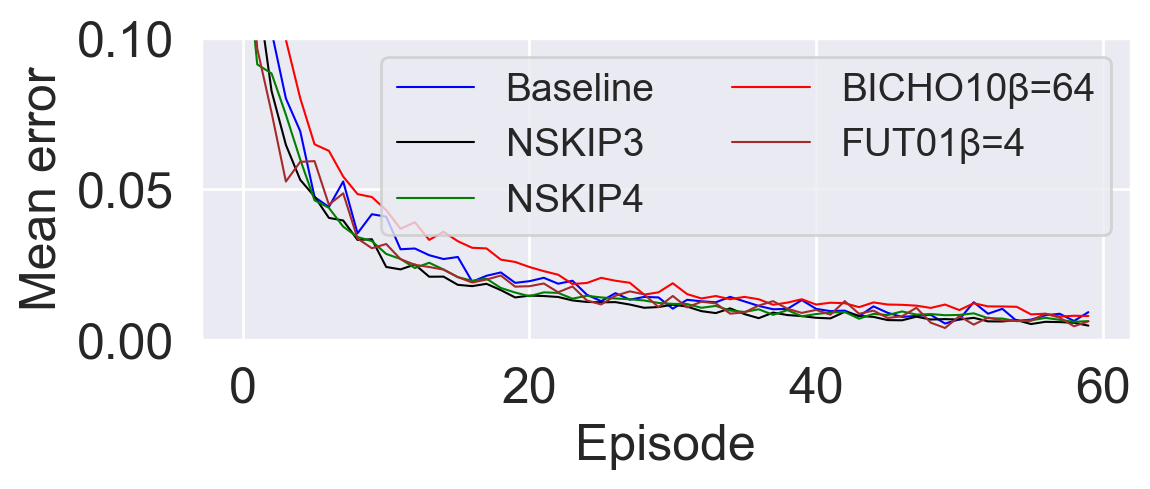}
  \label{fig:apx_Online_CP}
\end{minipage}
\hfill
\begin{minipage}{0.45\textwidth}
  \centering
  \includegraphics[width=\linewidth]{images/PU_ON_cum_max_reward_relative.png}
  \includegraphics[width=\linewidth]{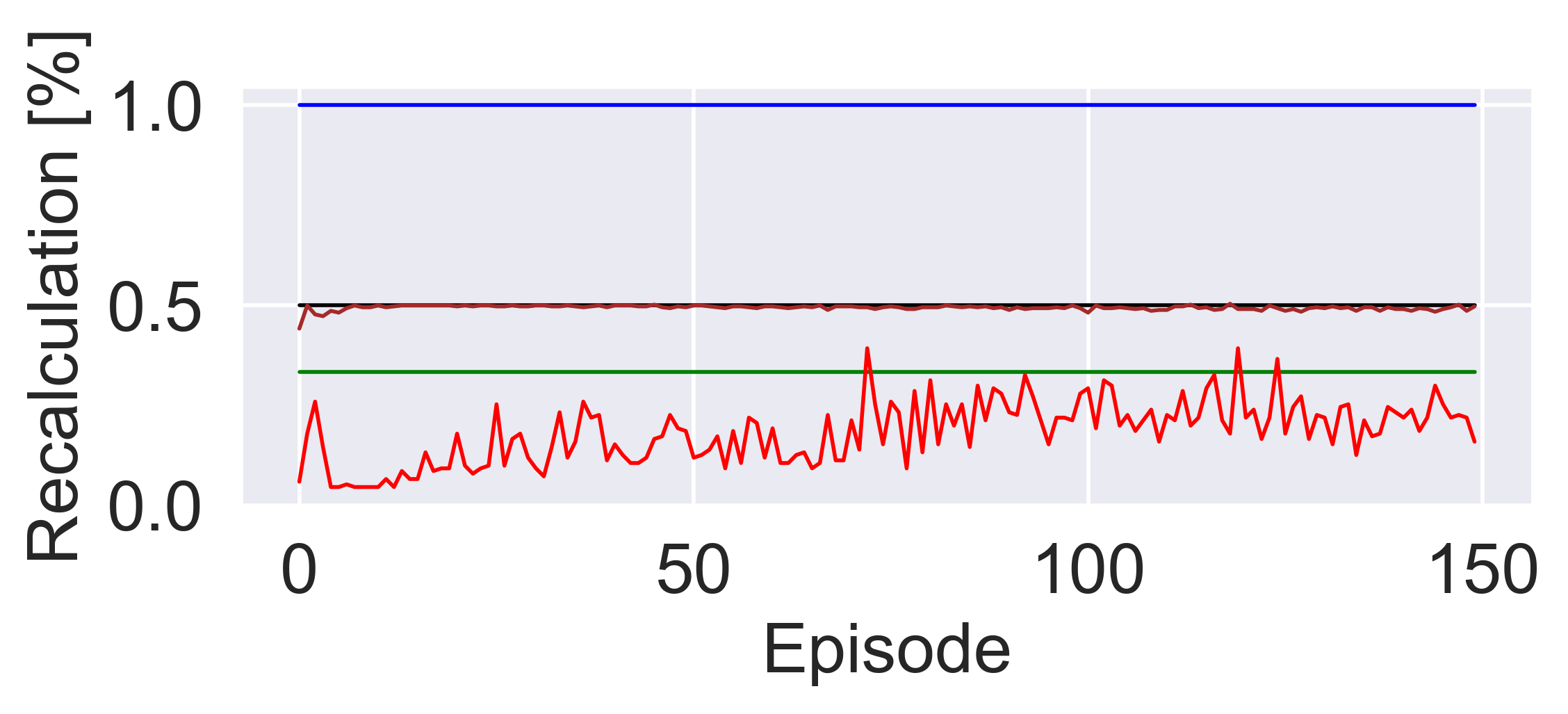}
  \includegraphics[width=\linewidth]{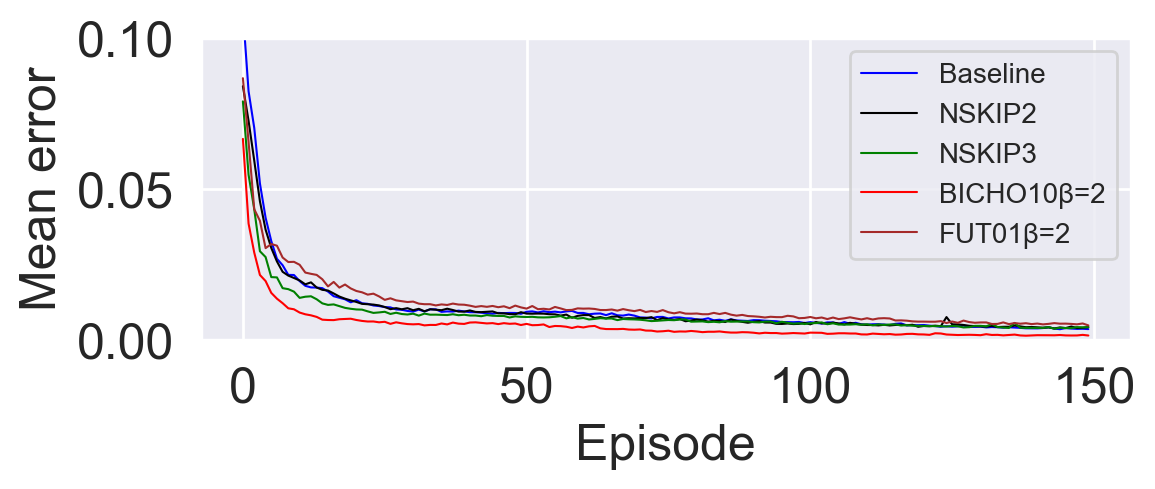}
  \label{fig:apx_Online_PU}
\end{minipage}
\caption{Performance of different methods while training the dynamics model with skipping. Top) Episode reward (y-axis) in relation to the number of the relative wall time (x-axis). The vertical lines represent the end of training for the method in question. Middle) Percentage of replanning steps (y-axis) in relation to the number of training iterations (x-axis). Bottom) Mean error of the observations predicted by the dynamics model and the observation from the environment. Left) Results in the CP environment. Right) Results in the PU environment.}
\label{fig:apx_Online_CP_PU}
\end{figure}

\begin{figure}
\centering
\begin{minipage}{0.45\textwidth}
  \centering
  \includegraphics[width=\linewidth]{images/RE_ON_cum_max_reward_relative.png}
  \includegraphics[width=\linewidth]{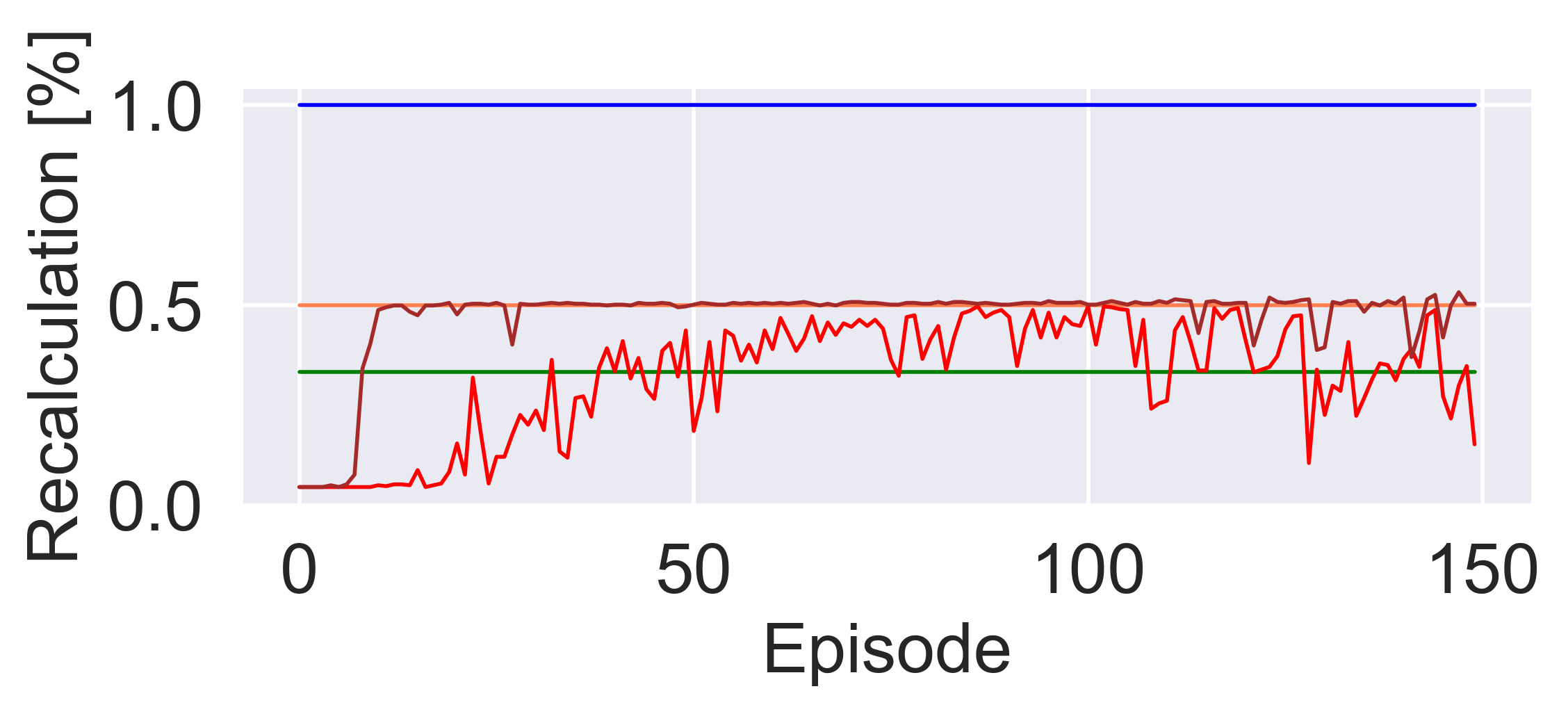}
  \includegraphics[width=\linewidth]{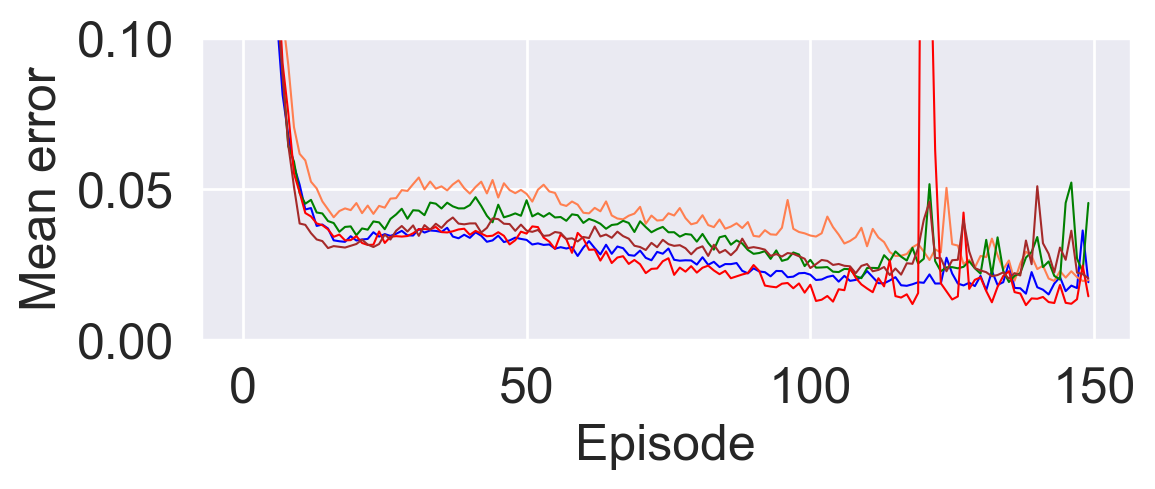}
  \label{fig:apx_Online_RE}
\end{minipage}
\hfill
\begin{minipage}{0.45\textwidth}
  \centering
  \includegraphics[width=\linewidth]{images/HC_ON_cum_max_reward_relative.png}
  \includegraphics[width=\linewidth]{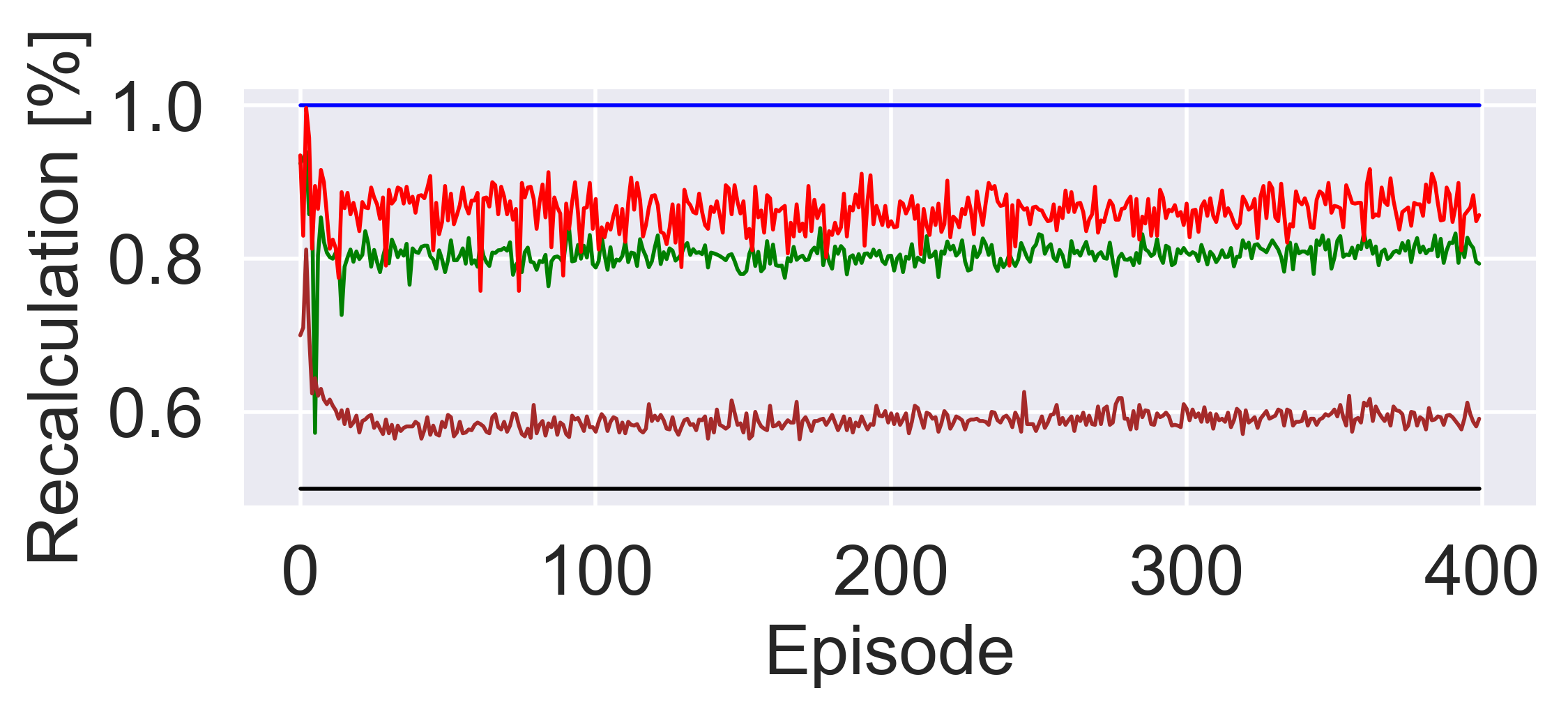}
  \includegraphics[width=\linewidth]{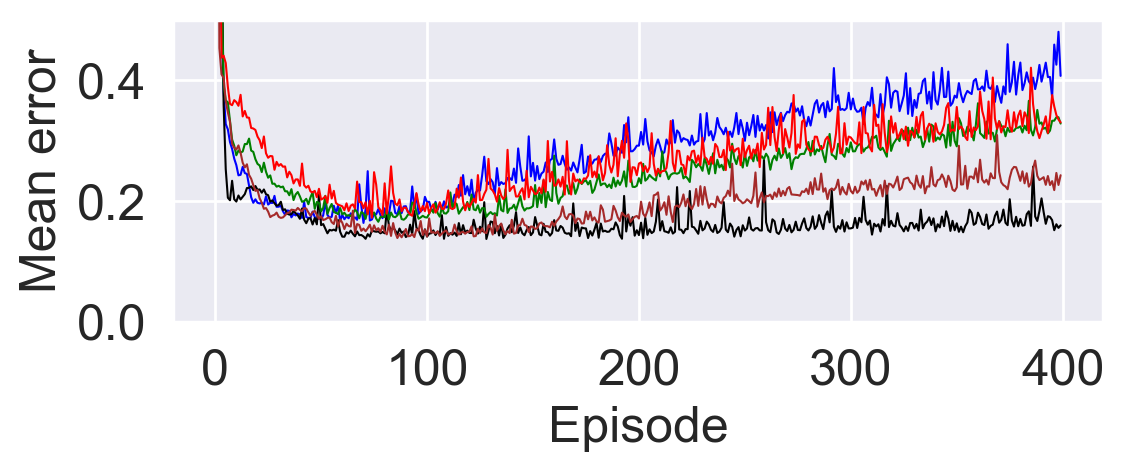}
  \label{fig:apx_Online_HC}
\end{minipage}
\caption{Performance of different methods while training the dynamics model with skipping. Top) Episode reward (y-axis) in relation to the number of the relative wall time (x-axis). The vertical lines represent the end of training for the method in question. Middle) Percentage of replanning steps (y-axis) in relation to the number of training iterations (x-axis). Bottom) Mean error of the observations predicted by the dynamics model and the observation from the environment. Left) Results in the RE environment. Right) Results in the HC environment.}
\label{fig:apx_Online_RE_HC}
\end{figure}

Figure \ref{fig:apx_Online_CP_PU} and \ref{fig:apx_Online_RE_HC} show the resulting performance of different hyper parameters while training the dynamics model with skipping in CP, PU, RE and HC. Table \ref{tbl:online_results} shows numerical results for each environment. CP was trained for 60 episodes, RE and PU for 150 and finally HC for 400. Rw represents the average over 3 experiments of the maximum rewards seen so far, RcPer\@Max is the percentage of replanning steps when the algorithm reached the maximum Rw. EpNr$@$Max number of episodes needed to reach Rw and RelEp\#$@$Max represents the relative wall time compared to the baseline when the algorithm reached the maximum Rw.



 \clearpage

\begin{table}[h]
\centering
\begin{tabular}{@{}lllll@{}}
\toprule
\multicolumn{5}{c}{CP}\\ \midrule
Method  & \multicolumn{1}{c}{Rw} & \multicolumn{1}{c}{RcPer$@$Max}  & \multicolumn{1}{c}{EpN$@$Max} & \multicolumn{1}{c}{RelEp\#$@$Max} \\
\midrule
Baseline	 & 181.6799	 & 1.0000	 & 57	 &    58.00\\
NSKIP3	 & 180.7760	 & 0.3350	 & 59	 &    20.10\\
NSKIP4	 & 180.6540	 & 0.2500	 & 58	 &    14.75\\
BICHO10$\beta$64	 & 180.8982	 & 0.1475	 & 58	 &     6.72\\
FUT01$\beta$4	 & 180.1486	 & 0.2567	 & 47	 &    10.53\\
\midrule

\multicolumn{5}{c}{PU}  \\
\midrule
Baseline	 & -47.9417	 & 1.0000	 & 142	 &   143.00\\
NSKIP2	 & -52.0129	 & 0.5000	 & 143	 &    72.00\\
NSKIP3	 & -57.3462	 & 0.3333	 & 147	 &    49.33\\
BICHO10$\beta$2	 & -49.0292	 & 0.1733	 & 136	 &    24.61\\
FUT01$\beta$2	 & -52.8447	 & 0.4956	 & 135	 &    67.31\\

\midrule
\multicolumn{5}{c}{RE}  \\
\midrule
Baseline	 & -33.5717	 & 1.0000	 & 70	 &    71.00\\
NSKIP2	     & -34.4023	 & 0.5000	 & 82	 &    41.50\\
NSKIP3	     & -34.2190	 & 0.3333	 & 72	 &    24.33\\
BICHO10$\beta$512	 & -37.7357	 & 0.4822	 & 95	 &   28.08\\
FUT01$\beta$16	 & -35.7938	 & 0.1333	 & 91	 &    12.64\\

\midrule
\multicolumn{5}{c}{HC}  \\
\midrule
Baseline	 & 22491.9876	 & 1.0000	 & 372	 &   373.00\\
NSKIP2	 & 6266.1845	 & 0.5000	 & 77	 &    39.00\\
BICHO05$\beta$=0.05	 & 18323.8463	 & 0.8305	 & 384	 &   310.31\\
BICHO10$\beta$=0.05	 & 20787.0499	 & 0.8640	 & 294	 &   254.11\\
FUT01$\beta$=0.1	 & 12605.5204	 & 0.5830	 & 259	 &   153.11\\

\midrule

\end{tabular}
\caption{Performance of each method on the CP, PU, RE and HC environment while updating the dynamics model online. Rw: average over 3 experiments of the maximum rewards seen so far. RcPer$@$Max: percentage of replanning steps when the algorithm reached the maximum Rw. EpNr$@$Max number of episodes needed to reach Rw. RelEp\#$@$Max: relative wall time when the algorithm reached the maximum Rw.}
\label{tbl:online_results}
\end{table}
\clearpage